\documentclass[runningheads]{llncs}
\usepackage[T1]{fontenc}
\usepackage{graphicx}
\usepackage{booktabs}
\usepackage[misc]{ifsym}

\usepackage{amsmath,amsfonts,bm}

\def\eqref#1{equation~\ref{#1}}
\def\Eqref#1{Equation~\ref{#1}}

\def\1{\bm{1}}

\DeclareMathAlphabet{\mathsfit}{\encodingdefault}{\sfdefault}{m}{sl}
\SetMathAlphabet{\mathsfit}{bold}{\encodingdefault}{\sfdefault}{bx}{n}

\DeclareMathOperator*{\argmax}{arg\,max}
\DeclareMathOperator*{\argmin}{arg\,min}

\usepackage{mwe}
\usepackage{hyperref}
\hypersetup{hidelinks}
\usepackage{url}
\usepackage{subcaption}
\usepackage{multicol}
\usepackage{bm}
\usepackage{multirow}
\usepackage{longtable}
\usepackage{algorithm}
\usepackage{algpseudocode}
\usepackage{array}
\usepackage{wrapfig}
\usepackage[figuresleft]{rotating}
\makeatletter
\newcommand{\res}[2]{
  \def\temp@series{\f@series}
  \ifx\temp@series\text@bold@series
    \boldmath
  \else
    \if b\expandafter\@car\f@series\@nil
      \boldmath
    \fi
  \fi

  $#1_{( \text{\scriptsize #2}) }$
}
\makeatother

\begin{document}

\title{Constraint-Aware Discrete Black-Box Optimization Using Tensor Decomposition\texorpdfstring{\thanks{This author-prepared preprint is based on the originally submitted manuscript, with minor typographical corrections. The Version of Record of this contribution is published in \emph{Machine Learning and Knowledge Discovery in Databases. Research Track}, and is available online at \url{https://doi.org/10.1007/978-3-032-37667-1_35}.}}{}}
\titlerunning{Constraint-Aware Discrete BBO Using Tensor Decomposition}

\author{
Keisuke~Onoue\inst{1}\orcidID{0009-0007-3249-6572}\\
and Ryosuke~Kojima\inst{2,3}\orcidID{0000-0003-1095-8864} (\Letter)
}

\authorrunning{K. Onoue and R. Kojima}

\institute{Nara Institute of Science and Technology, Nara, Japan \email{onoue.keisuke.ok2@naist.ac.jp}
\and
Graduate School of Medicine, Kyoto University, Kyoto, Japan \email{kojima.ryosuke.8e@kyoto-u.ac.jp}
\and
RIKEN Center for Biosystems Dynamics Research, Japan }

\maketitle
\setcounter{footnote}{0}

\begin{abstract}
Discrete black-box optimization is often addressed using approaches such as Sequential Model-Based Optimization (SMBO), which aims to improve sample efficiency by fitting surrogate models that approximate a costly objective function over a discrete search space.
In many real-world problems, the set of feasible inputs is often given by logical constraints known in advance.
However, existing surrogate modeling techniques generally fail to capture the symbolic rules governing feasibility in discrete input spaces.
In this paper, we propose a surrogate modeling approach based on tensor decomposition that captures the structure of discrete search spaces while directly integrating feasibility information.
To implement this approach, we formulate surrogate model training as a constrained polynomial optimization problem and solve a relaxed formulation using a differentiable penalty term derived from T-norms.
Our experiments on both synthetic and real-world benchmarks, including a pressure vessel design task, demonstrate that the proposed method improves sample efficiency by effectively guiding the search away from infeasible regions.
\end{abstract}

\section{Introduction}
\label{sec:introduction}

\begin{figure}[t]
  \centering
  \includegraphics[width=0.60\textwidth]{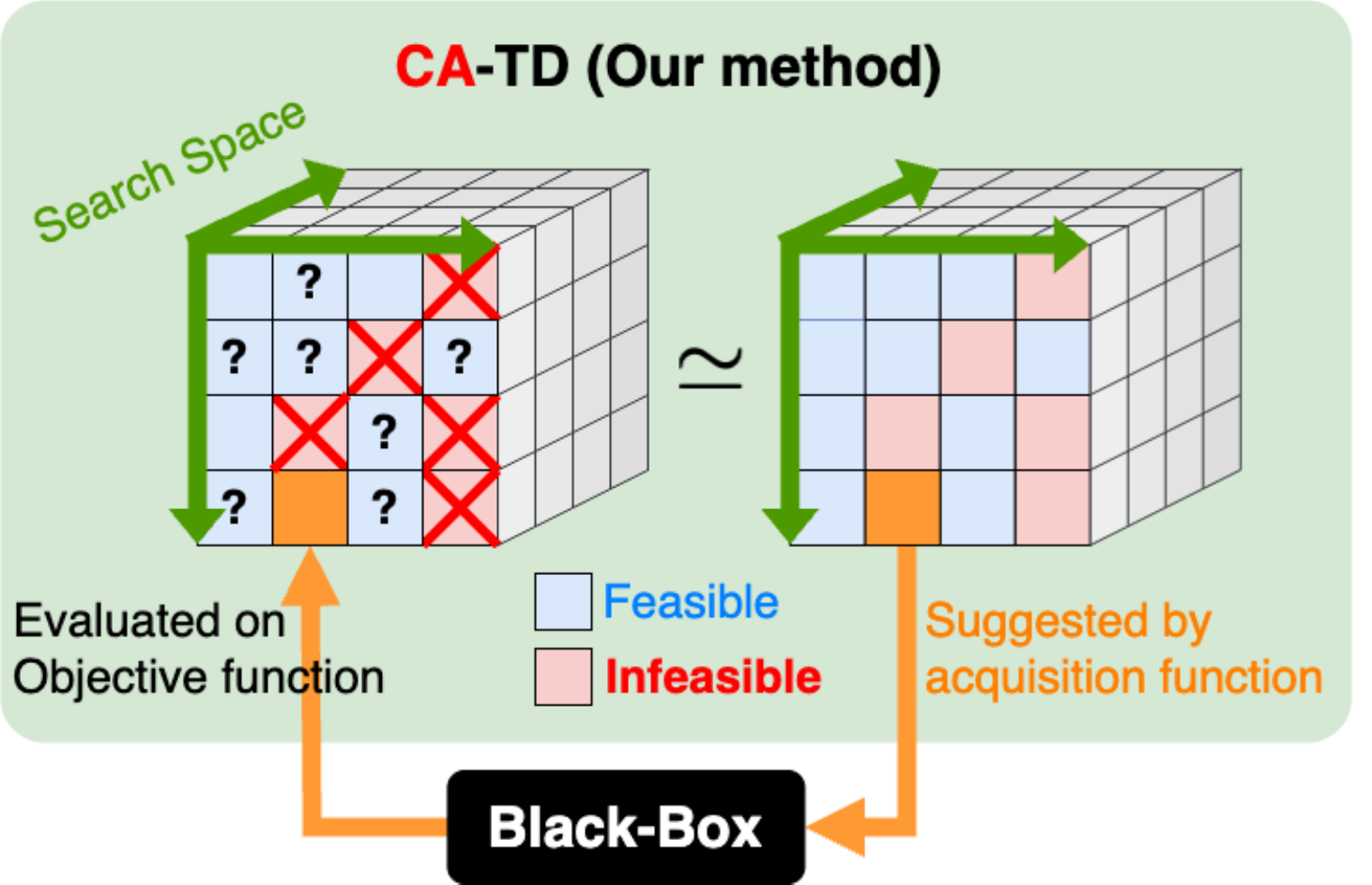}
  \caption{Overview of CA-TD: CA-TD approximates the black-box objective on the feasible region, effectively guiding the search in SMBO.}
  \label{fig:our_method}
\end{figure}

Black-box optimization (BBO) aims to find optimal inputs for an objective function that can only be accessed through input-output data~\cite{rios2013derivative} and has been widely used in fields such as engineering design~\cite{coello2002constraint}, material discovery~\cite{frazier2016materialdesign}, and hyperparameter tuning for machine learning~\cite{he2021automl,bergstra2011hyperparameter}.
Because evaluating such objective functions is often costly in terms of monetary cost, execution time, and computational resources, sample-efficient frameworks such as Sequential Model-Based Optimization (SMBO) have been developed~\cite{hutter2011sequential,shahriari2015humanout}.
SMBO uses a surrogate model to approximate the objective function and an acquisition function to balance exploration and exploitation when choosing new samples.

This paper considers BBO over discrete search spaces that are commonly encountered in real-world applications, such as categorical parameters representing component choices in engineering design~\cite{papalexopoulos2022constrained,gonzalez2024survey,zamuda2018black}.
For such discrete BBO problems, methods based on tensor decomposition (TD) offer a sample-efficient approach that has recently demonstrated strong potential~\cite{sozykin2022ttopt,chertkov2022optimatt,batsheva2023protes}.
In this approach, a discrete search space is represented by a tensor, and a surrogate model is constructed by approximating this tensor, for example, with a low-rank tensor.

For many real-world problems addressed by BBO, input constraints arising from safety requirements, manufacturing capabilities, or design rules are crucial.
A typical way to incorporate such input constraints into SMBO is to evaluate feasibility when acquiring new samples, for example by rejecting infeasible inputs or modifying the acquisition function~\cite{gardner2014bayesian,gelbart2014bayesian}.
More advanced methods also follow this paradigm, employing sophisticated solvers to optimize the acquisition function over the known feasible domain~\cite{papalexopoulos2022constrained}. 
In these approaches, however, the surrogate model itself is typically learned as a purely data-driven black-box, ignoring the underlying symbolic structure of the input space. 
Disregarding these logical constraints leads to poor approximation in sparse feasible regions, as the model must learn complex boundaries purely from a limited number of objective evaluations.
To address this, we propose embedding symbolic domain knowledge directly into the surrogate model's architecture. 
By making the model constraint-aware during training, we effectively integrate discrete logical constraints directly into the function approximation process.
This integration enhances sample efficiency in restricted search spaces.

In this paper, we propose a logic-integrated surrogate modeling approach named Constraint-Aware Tensor Decomposition (CA-TD). 
Our method represents the objective function using TD while incorporating logical constraints given a priori.
We formulate surrogate learning as a constrained polynomial optimization problem (POP), where logical constraints are interpreted as requirements on surrogate tensor entries.
To ensure scalability, we leverage differentiable logic~\cite{badreddine2022logic,xu2018semantic} by employing T-norms to relax these discrete symbolic rules into a differentiable penalty function. 
This framework enables the integration of logical domain knowledge into a gradient-based learning process.

Our contribution is threefold:
\begin{itemize}
    \item We formulate the training of a TD-based surrogate model with logical input constraints as a POP, defining the CA-TD framework.
    \item We develop a gradient-based training method for CA-TD by introducing a T-norm-based differentiable penalty term, thereby enabling the method to scale to large discrete search spaces.
    \item We evaluate our approach on a diverse set of synthetic and real-world benchmarks, including a classic engineering design task. The results demonstrate improvements in sample efficiency compared to conventional methods.
\end{itemize}
Our source code is publicly available\footnote{\url{https://github.com/k-onoue/c-bbo}}.

\section{Preliminaries}
\label{sec:preliminaries}

This section introduces the fundamental concepts underlying our proposed method.
Because our problem formulation is based on the discrete BBO approach within the SMBO framework, we begin by providing an overview of BBO and SMBO.
Additionally, we describe TD-based surrogate models that our method employs, focusing in particular on tensor train (TT) decomposition.

\subsection{Discrete Black-Box Optimization Problem and Sequential Model-Based Optimization}
\label{ssec:bbo}

\begin{wrapfigure}{r}{0.5\textwidth}
    \vspace{-46pt}
    \begin{minipage}{\linewidth}
        \begin{algorithm}[H]
            \caption{SMBO}
            \label{alg:smbo_prelim}
            \begin{algorithmic}[1]
                \Require Objective $g$, Search space $\mathcal{X}$, Surrogate $f$, Acquisition $\alpha$, Iteration $T$
                \State Initialize $\mathcal{H} = \{\}$
                \For{$t = 1$ to $T$}
                    \State Update $f_{t-1}$ using $\mathcal{H}$ 
                    \State $\mathbf{x}_t \gets \argmax_{\mathbf{x} \in \mathcal{X}} \alpha(\mathbf{x}, f_{t-1})$
                    \State $y_t \gets g(\mathbf{x}_t)$
                    \State $\mathcal{H} \gets \mathcal{H} \cup \{(\mathbf{x}_t, y_t)\}$
                \EndFor
                \State \Return $\mathcal{H}$
            \end{algorithmic}
        \end{algorithm}
    \end{minipage}
    \vspace{-10pt}
\end{wrapfigure}

First, we formulate the discrete BBO problem that underlies our setting.
Consider a search space $\mathcal{X} = X_1 \times \cdots \times X_d$, where $X_k$ is a finite set for $k=1,\dots,d$. 
Let $g: \mathcal{X} \to \mathbb{R}$ be an objective function.

The goal of this problem is to find
\[
    \mathbf{x}^\star \;=\; \argmin_{\mathbf{x}\in\mathcal{X}} g(\mathbf{x}).
\]
In this problem, no further information about $g$ is available, such as its derivative, so it is called a black-box function.
For each iteration $t$, we denote the evaluated objective value as $y_t = g(\mathbf{x}_t)$ and the accumulated set of observations as the history $\mathcal{H} = \{(\mathbf{x}_i, y_i)\}_{i=1}^t$.
In practice, it is assumed that evaluating $g$ is costly, and it is desirable to obtain a good solution with as few evaluations of $g$ as possible.

The overall SMBO procedure is summarized in Algorithm~\ref{alg:smbo_prelim}. 
SMBO is a general framework of BBO that includes Bayesian optimization as a special case, and we adopt a variant commonly used in Bayesian optimization~\cite{frazier2018tutorial,shahriari2015humanout,bergstra2011hyperparameter}, which iteratively performs the following three steps.
1) A probabilistic surrogate model $f$ is fitted to all previous observations.
Instead of directly evaluating the costly function $g$, the surrogate model $f$ is used to approximate $g$.
2) The most promising point to evaluate next is selected using an acquisition function $\alpha(\cdot)$, defined based on the surrogate function $f$.
In our implementation, we utilize the Expected Improvement (EI) criterion~\cite{Mockus1978ei} as the acquisition function to decide the next point.
3) The objective function $g$ at the selected point is evaluated.
This loop is repeated $T$ times. 

\subsection{Tensor Decomposition Surrogate Models}
\label{ssec:td}

This section describes TD-based surrogate models for approximating black-box functions $g$.
We define tensor contraction as the fundamental operation for our surrogate model. 
Given two tensors $\mathbf{A} \in \mathbb{R}^{n_1 \times \dots \times n_p \times r}$ and $\mathbf{B} \in \mathbb{R}^{r \times m_1 \times \dots \times m_q}$, their contraction over the common index of size $r$ is defined as a tensor of order $p+q$ whose entries are:
\begin{equation}
(\mathbf{A}\mathbf{B})[i_1, \dots, i_p, j_1, \dots, j_q] := \sum_{k=1}^r \mathbf{A}[i_1, \dots, i_p, k] \mathbf{B}[k, j_1, \dots, j_q].
\end{equation}

Before considering the surrogate model, we represent the observed objective values as a partially observed tensor $\mathcal{Y} \in \mathbb{R}^{|X_1| \times \cdots \times |X_d|}$, where only entries corresponding to evaluated points in $\mathcal{H}$ are known.
Each entry $\mathcal{Y}[\mathbf{x}]$ represents the objective value $g(\mathbf{x})$.
A TD-based surrogate $\hat{\mathcal{Y}}$ approximates $\mathcal{Y}$ using a low-rank structure. We consider three formats, Canonical Polyadic (CP), Tensor Ring (TR) and Tensor Train (TT), where $R$ denotes the rank (complexity) parameter:

\paragraph{Canonical Polyadic Decomposition:} 
CP decomposition~\cite{kolda2009cp} models a tensor as a sum of $R$ rank-one tensors:
\begin{equation*}
    \hat{\mathcal{Y}}[x_1, \dots, x_d] = \sum_{r=1}^{R} \mathbf{U}^{(1)}[x_1, r] \cdot \mathbf{U}^{(2)}[x_2, r] \cdots \mathbf{U}^{(d)}[x_d, r]
\end{equation*}
where $\mathbf{U}^{(k)} \in \mathbb{R}^{|X_k| \times R}$ are factor matrices.

\paragraph{Tensor Ring Decomposition:} 
TR format~\cite{zhao2016ring} represents a tensor as a circular product of third-order cores $\mathcal{G}^{(k)} \in \mathbb{R}^{r_{k-1} \times |X_k| \times r_k}$:
\begin{equation*}
    \hat{\mathcal{Y}}[x_1, \dots, x_d] = \text{Tr}(\mathbf{G}^{(1)}[x_1] \mathbf{G}^{(2)}[x_2] \cdots \mathbf{G}^{(d)}[x_d])
\end{equation*}
where $\text{Tr}(M) := \sum_{i} M_{ii}$. 
For TR, $r_0 = r_d$ is required to satisfy the trace operation.
For simplicity, we adopt a uniform rank setting where all ranks are fixed to $R$ (i.e., $r_k = R$ for all $k \in \{0, \dots, d\}$).

\paragraph{Tensor Train Decomposition:} 
TT decomposition is a special case of TR where boundary ranks are set to one ($r_0 = r_d = 1$), breaking the circular connection.
Here, $\mathbf{G}^{(k)}[x_k] \in \mathbb{R}^{r_{k-1} \times r_k}$ denotes the $x_k$-th frontal slice of the third-order core tensor $\mathcal{G}^{(k)} \in \mathbb{R}^{r_{k-1} \times |X_k| \times r_k}$. 
Then we have
\begin{equation}
    \hat{\mathcal{Y}}[\mathbf{x}]
    \;=\;
    \mathbf{G}^{(1)}[x_1]
    \,\mathbf{G}^{(2)}[x_2]\,
    \cdots\,
    \mathbf{G}^{(d)}[x_d],
    \label{eq:tt_entry}
\end{equation}
where $\mathbf{G}^{(k)}[x_k] \in \mathbb{R}^{r_{k-1} \times r_k}$. 
For the internal ranks, we use a uniform setting $r_k = R$ for all $k \in \{1, \dots, d-1\}$. 
The number of parameters scales as $\mathcal{O}(d\,n\,R^2)$, where $n := \max_k |X_k|$. 

In TD, the surrogate is trained to minimize the mean squared error $\mathcal{L}_{\text{recon}}$ on the observation history $\mathcal{H} = \{(\mathbf{x}_i, y_i)\}_{i=1}^t$:
\begin{equation}
    \mathcal{L}_{\text{recon}}
    \;=\;
    \frac{1}{|\mathcal{H}|} \sum_{\mathbf{x} \in \mathcal{H}} \left( \mathcal{Y}[\mathbf{x}] - \hat{\mathcal{Y}}[\mathbf{x}] \right)^2.
    \label{eq:tt_loss}
\end{equation}

\subsection{T-norms and Differentiable Logic}
In neuro-symbolic AI~\cite{hitzler2022neural}, T-norms are used to generalize classical logic to a continuous domain, enabling the integration of symbolic rules into gradient-based learning. 
A T-norm $T:[0,1]^2 \to [0,1]$ models logical conjunction ($\wedge$). Its dual, a T-conorm $S$, models disjunction ($\vee$). 
For instance, the {\L}ukasiewicz T-norm is defined as $T_L(a, b) = \max(0, a + b - 1)$.
Using these norms, a logical implication $A \implies B$ can be relaxed into a differentiable function using the residuum-based implication: $I(a, b) = \sup \{ z \in [0,1] \mid T(a, z) \leq b \}$. For the {\L}ukasiewicz norm, this results in $I_L(a, b) = \min(1, 1 - a + b)$. 
We utilize this framework to translate discrete constraints into differentiable objectives for our surrogate model.

\section{Proposed Method: Constraint-Aware Tensor Decomposition Surrogate}
\label{sec:method}

Our method integrates constraint-awareness directly into the learning process of the TD-based surrogate model for SMBO.

\subsection{Problem Formulation}
\label{sec:problem_formulation}

First, we describe our formulation of the discrete BBO under logical input constraints.
Given a discrete search space $\mathcal{X} = X_1 \times \cdots \times X_d$ and an objective function $g: \mathcal{X} \to \mathbb{R}$, the goal is to find
\begin{equation*}
    \mathbf{x}^\star
    \;=\;
    \argmin_{\mathbf{x} \in \mathcal{X}} g(\mathbf{x})
    \quad \text{subject to} \quad c(\mathbf{x}) = 1,
\end{equation*}
with a constraint function $c: \mathcal{X} \to \{0,1\}$ whose evaluation cost is 
negligible compared to $g$.
Here, we introduce the notation $\mathcal{X}_{\mathrm{feas}} := \{ \mathbf{x} \in \mathcal{X} \mid c(\mathbf{x}) = 1 \}$ and rewrite the above problem as $\mathbf{x}^\star = \argmin_{\mathbf{x} \in \mathcal{X}_{\mathrm{feas}}} g(\mathbf{x})$ for simplicity.

\subsection{A Formulation as a Polynomial Optimization Problem}
\label{ssec:pop_formulation}

We incorporate input constraints into the surrogate model. 
Specifically, we assume that evaluations at infeasible inputs yield objective values at least as high as the maximum feasible value observed so far.
Formally, at each iteration $t$, we define the threshold $\tau_t$ as:
\begin{equation*}
\tau_t := \max \{ y \mid (\mathbf{x}, y) \in \mathcal{H}_t \text{ and } c(\mathbf{x}) = 1 \}.
\end{equation*}
If no feasible points have been observed, $\tau_t$ is set to a predefined sufficiently large value. 
The surrogate model $\hat{\mathcal{Y}}$ is then required to satisfy $\hat{\mathcal{Y}}[\mathbf{x}] \geq \tau_t$ for all $\mathbf{x} \in \mathcal{X}_{\mathrm{infeas}}$.
Thus, the surrogate learning under input constraints is formulated as follows:
\begin{equation}
\label{eq:pop}
\begin{aligned}
    &\min_{\hat{\mathcal{Y}}} \quad
    \frac{1}{|\mathcal{H}|}
    \sum_{\mathbf{x} \in \mathcal{H}}
    \left( \mathcal{Y}[\mathbf{x}] - \hat{\mathcal{Y}}[\mathbf{x}] \right)^2, \\
    &\text{subject to} \quad
    \hat{\mathcal{Y}}[\mathbf{x}] \geq \tau_t
    \quad \text{for all} \quad
    \mathbf{x} \in \mathcal{X}_{\mathrm{infeas}}.
\end{aligned}
\end{equation}

Because the surrogate tensor $\hat{\mathcal{Y}}[\mathbf{x}]$ is a polynomial function of the core tensor parameters according to \Eqref{eq:tt_entry}, the above problem constitutes a POP (Appendix~\ref{app:pop}).
An established approach to solving POPs involves constructing a hierarchy of semidefinite programming (SDP) relaxations~\cite{lasserre2001global}, which can provide arbitrarily tight lower bounds on the global optimum.
We apply this approach to the problem in \Eqref{eq:pop} and simply refer to the resulting method as HSDP hereafter.

\subsection{Penalized Loss Function}
\label{ssec:penalty_loss}

To ensure scalability, we leverage differentiable logic to integrate discrete prior knowledge into the continuous learning objective. Specifically, we interpret the feasibility requirement as a logical rule: for any input $\mathbf{x}$, if $\mathbf{x} \in \mathcal{X}_{\mathrm{infeas}}$, then the surrogate output $\hat{\mathcal{Y}}[\mathbf{x}]$ should be at least the current threshold $\tau_t$.

Following the principles of differentiable logic~\cite{badreddine2022logic,xu2018semantic}, we relax this discrete rule using T-norms, which map logical truth values to the interval $[0, 1]$. 
We utilize the {\L}ukasiewicz logic framework, where the truth value of an implication $A \implies B$ is defined by the residuum $I(a, b) = \min(1, 1 - a + b)$. 
In our context, for an infeasible point where the antecedent is treated as true (i.e., it is strictly infeasible), the degree of violation of the requirement $\hat{\mathcal{Y}}[\mathbf{x}] \geq \tau_t$ is given by $1 - I(1, s(\mathbf{x}))$, where $s(\mathbf{x})$ is a score representing the truth of the condition. 
By mapping the surrogate's output to the threshold, this logical violation is directly translated into the hinge loss: $\max(0, \tau_t - \hat{\mathcal{Y}}[\mathbf{x}])$.

We train the TD-surrogate by minimizing a total loss function that balances empirical data fitting with this logical penalty:
\begin{equation}
\mathcal{L}_{\mathrm{total}} = \mathcal{L}_{\mathrm{recon}} + \lambda \cdot \mathcal{L}_{\mathrm{pen}}.
\label{eq:pen_loss_total}
\end{equation}
The reconstruction loss $\mathcal{L}_{\mathrm{recon}}$ is defined in \Eqref{eq:tt_loss}. The penalty term $\mathcal{L}_{\mathrm{pen}}$ is defined as the expected violation over the infeasible region:
\begin{equation*}
\mathcal{L}_{\mathrm{pen}} = \mathbb{E}_{\mathbf{x} \sim \mathcal{U}(\mathcal{X}_{\mathrm{infeas}})} \left[ \max(0, \tau_t - \hat{\mathcal{Y}}[\mathbf{x}]) \right],
\end{equation*}
where $\mathcal{U}(\mathcal{X}_{\mathrm{infeas}})$ is a uniform distribution. For small search spaces, this expectation is computed exactly, while for large spaces, it is estimated via mini-batch sampling of infeasible points.

This penalty term enforces the logical boundary by discouraging the surrogate from assigning low values to regions known to be infeasible. 
We refer to this strategy of embedding logical constraints via gradient-based optimization as PGRAD (Penalty with Gradient-based optimization).

\subsection{Uncertainty Quantification for the Acquisition Function}
\label{ssec:ensemble_acquisition}

To balance exploration and exploitation in SMBO, we quantify predictive uncertainty using an ensemble of $M$ models $\{\hat{\mathcal{Y}}^{(m)}\}_{m=1}^M$. 
Each model is trained independently, starting from different random initializations of the TT core parameters. 

For any input $\mathbf{x}$, the set of ensemble predictions $\{ \hat{\mathcal{Y}}^{(m)}[\mathbf{x}] \}_{m=1}^M$ induces an empirical predictive distribution $Y(\mathbf{x})$.
We compute the sample mean $\mu(\mathbf{x})$ and standard deviation $\sigma(\mathbf{x})$ from this distribution to evaluate the EI acquisition function:
\begin{equation*}
    \alpha_\mathrm{EI}(\mathbf{x}) = \mathbb{E} \left[ \max(0, y^\star - Y(\mathbf{x})) \right],
\end{equation*}
where $y^\star$ is the best (minimum) feasible value observed so far. 
Here, the expectation is taken over the empirical distribution $Y(\mathbf{x})$. 
This ensemble-based approach allows the surrogate to guide the search by identifying regions with both low predicted values and high model uncertainty.

\section{Related Work}
\label{sec:related_work}

Our method builds on three areas of work: constrained BBO, tensor decomposition for black-box optimization (TD-BBO), and constrained tensor decomposition. 
We briefly review each area below and explain our unique contributions in relation to each field.

\subsection{Constrained Black-Box Optimization}
\label{ssec:cbo}

Constrained BBO deals with expensive objectives and expensive or inexpensive input constraints whose analytic forms are unknown~\cite{rios2013derivative}.  
Most existing algorithms extend Bayesian Optimization (BO)~\cite{frazier2018tutorial,shahriari2015humanout}, which is a form of SMBO that typically uses Gaussian Processes (GPs) as surrogate models~\cite{williams2006gaussian}.
Early GP-based approaches fit separate GPs to model each constraint and incorporate the estimated feasibility into the acquisition function, typically by combining them with EI~\cite{gardner2014bayesian,gelbart2014bayesian}.  
Subsequent work further extended this strategy using augmented Lagrangian methods~\cite{picheny2016bayesian} and level-set estimation techniques~\cite{zhang2023learning}.

Unlike the methods mentioned above, in the case of explicit hard constraints, some approaches maximize the acquisition function within the feasible region.
For example, a method that combines GPs with mixed-integer programming to maximize the acquisition function under known constraints has been proposed~\cite{thebelt2022tree}.
A more advanced method, NN+MILP, uses piecewise-linear neural networks with acquisition maximization via mixed-integer linear programming~\cite{papalexopoulos2022constrained}, allowing flexible integration of combinatorial constraints in discrete search spaces.

Most existing constrained BBO methods handle constraints by modifying the acquisition function, as mentioned above. 
Our proposed approach is distinguished by the direct incorporation of known feasibility information into the surrogate model training process. 
This is expected to encourage the surrogate model itself to learn the feasibility information, aiming for a more accurate approximation of the objective function.

\subsection{Logic Integration via Differentiable Logic}

Logic integration aims to combine continuous learning with symbolic reasoning by using T-norms to relax discrete rules into differentiable loss functions \cite{hitzler2022neural,badreddine2022logic}. 
While established frameworks like Logic Tensor Networks and Semantic Loss focus primarily on classification or reasoning tasks \cite{badreddine2022logic,xu2018semantic}, CA-TD extends this paradigm to black-box optimization. 
CA-TD bridges the gap between function approximation and symbolic knowledge by reformulating surrogate learning as a logic-constrained optimization problem. 
By leveraging \L{}ukasiewicz logic within a tensor decomposition framework, we transform discrete constraints into a differentiable penalty landscape. 
This allows the tensor cores to be optimized not only for data fitting but also for consistency with the problem logic, enabling the surrogate model to internalize feasible domain boundaries even in regions where observation data is scarce.

\subsection{Tensor Decomposition for BBO}
\label{ssec:tdbbo}

TD compactly represents multi-dimensional arrays and is well-suited for capturing discrete structures.
OptimaTT~\cite{chertkov2022optimatt} adopts the TT format for discrete, unconstrained BBO, while PROTES~\cite{batsheva2023protes} incorporates input constraints by encoding the input feasibility as a binary tensor and using its TT decomposition to guide surrogate initialization.  
Although TTOpt~\cite{sozykin2022ttopt} also uses the TT format, it is primarily designed for continuous optimization.  

In the context of TD-BBO, methods such as OptimaTT and TTOpt are primarily designed for unconstrained optimization. 
A pioneering approach that incorporates input constraints is the PROTES method introduced above,  which uses these to guide the initial exploration in the proxy initialization step. 
In contrast, focusing on the entire search process rather than just the initialization step, CA-TD incorporates the constraints into the update of the surrogate model at each optimization step.
This aims to improve sample efficiency by maintaining awareness of feasible regions throughout the search process.

\subsection{Tensor Decomposition under Constraints}
\label{ssec:constraint_td}

Previous work on constrained TD has focused on linear algebraic properties such as non-negativity or orthogonality for interpretability \cite{alexandrov2022nonnegative}. 
Our work departs from these by imposing point-wise logical constraints derived from the task's feasibility rules, representing a novel application of neuro-symbolic principles to tensor-based global optimization.

\section{Experiments}
\label{sec:experiments}

We conduct three experiments to evaluate the effectiveness of CA-TD in constrained black-box optimization on discrete domains.
The objectives of these experiments are:
(1) to compare the performance and scalability of our proposed constrained training strategies, HSDP and PGRAD;
(2) to evaluate the effectiveness of our constraint-aware approach using the default tensor format against conventional and advanced baseline methods; and
(3) to further verify its performance against a strong baseline on a wider range of complex tasks while analyzing the impact of different tensor formats (CP, TR, and TT).
\subsection{Benchmarks} 

\begin{wraptable}{r}{0.45\textwidth}
    \vspace{-25pt}
    \caption{The settings of Ackley on the grid used in experiments. $\ell$ determines the grid range $\{-\ell, \dots, \ell\}^2$, $r$ is the radius for the circular constraint, and $T$ is the number of evaluations in SMBO.}
    \label{tab:ackley_settings}
    \centering
    \small
    \setlength{\tabcolsep}{5pt}
    \begin{tabular}{lcccc}
        \toprule
        Experiment & Grid size & $\ell$ & $r$ & $T$ \\
        \midrule
        \multirow{3}{*}{Expt. 1} & $3 \times 3$   & 1  & 1   & 5   \\
                                & $5 \times 5$   & 2  & 2   & 15  \\
                                & $7 \times 7$   & 3  & 3   & 25  \\
        \midrule
        Expt. 2                  & $65 \times 65$ & 32 & 10  & 500 \\
        \bottomrule
    \end{tabular}
    \vspace{-23pt}
\end{wraptable}

This subsection briefly introduces the benchmark problems used to evaluate our method.
Detailed mathematical formulations for all problems are provided in Appendix~\ref{app:benchmarks}.

\paragraph{Ackley:}
We use the standard Ackley synthetic function~\cite{adorio2005ackley}, to which we apply a simple geometric constraint boundary.
The search space is an integer grid of \(\{-\ell, \ldots, \ell\}^2\), where \(\ell \in \mathbb{Z}^+\) defines the boundary of the discrete domain.
The feasible region is defined by the circular constraint \(x_1^2 + x_2^2 \leq r^2\), where \(r \in \mathbb{R}^+\) is the radius.
Typically, we set \(r \leq \ell\) to ensure the existence of an infeasible region within the grid.
We use this problem to evaluate performance across different scales, with the specific settings for Experiment 1 and 2 detailed in Table~\ref{tab:ackley_settings}.

\paragraph{Pressure Vessel:}
This is a classic mixed-variable engineering design problem where the goal is to minimize manufacturing cost under physical constraints~\cite{coello2002constraint}.
We adapt this problem to our discrete setting by discretizing the two continuous variables into 10 uniform levels.

\paragraph{Warcraft:}
This benchmark is a grid-based path optimization problem~\cite{ahmed2022warcraft}, where the goal is to find an optimal path on a map with combinatorial constraints defining path validity.
We evaluate this problem on two different map sizes: 
a $2 \times 2$ grid with $7^4$ candidate paths and a $2 \times 3$ grid with $7^6$ candidate paths.

\paragraph{Diabetes:}
This is a real-world inspired task where constraints are derived from domain knowledge to find actionable and medically plausible treatment plans from patient data~\cite{smith1988using}.

\paragraph{Additional Tasks for Extensive Comparison:}
To deeply evaluate our approach against advanced baselines, we incorporate seven additional challenging tasks derived from \cite{papalexopoulos2022constrained}. 
These span diverse domains including the Generalized Assignment Problem (GAP A/B), Constrained Ising Models (Ising A/B), Neural Architecture Search (NAS SSS/TSS), and DNA Binding optimization (TfBind). 
The detailed formulations of all tasks are provided in Appendix~\ref{app:benchmarks}.

\subsection{Experimental Setup}
\label{ssec:exp_setup}

Since the TT format is widely adopted in recent tensor-based BBO literature (e.g., OptimaTT, TTOpt, PROTES), we primarily use TT as the default format for our CA-TD models in the main baseline comparisons. 
However, because our constraint-aware framework is agnostic to the underlying tensor format, we also comprehensively investigate the performance of CP and TR formats in later experiments.

Each run uses a fixed evaluation  (see Table~\ref{tab:ackley_settings}), initialized from a random feasible input.
The ensemble size to quantify the uncertainty for TD-based surrogate models is set to $M=10$.
In Experiment~1, the surrogate mean is used for the acquisition function for simplicity, and EI is applied in Experiment~2.
The metrics include the best value obtained from the objective function and the round in which this value first appeared.
All results are averaged over 10 seeds.

HSDP is implemented using the Ncpol2sdpa~\cite{wittek2015algorithm} package with relaxation order 2.
The generated SDP problems are solved using a sparse semidefinite programming solver.

PGRAD uses the Adam optimizer~\cite{kingma2014adam} to minimize the total loss (\Eqref{eq:pen_loss_total}).  
The surrogate tensor is normalized to the range $[0, 1]$, and the penalty coefficient is fixed at $\lambda = 1$.
An ablation study on different values of $\lambda$ is provided in  Appendix~\ref{app:ablation_lambda}.
Training continues across SMBO rounds without reinitializing the tensor cores.  
This minimization at each round is terminated either when the loss drops below 0.1 or after 1000 epochs, whichever occurs first.
The detailed implementation, including software and hardware, is provided in Appendix~\ref{app:reproducibility}.

For comparison, we consider four baseline methods, each with an unconstrained (-u) and a constrained (-c) variant.
As typical BBO methods, we use Bayesian optimization based on Gaussian process (GP-u) and the Tree-structured Parzen Estimator (TPE-u)~\cite{watanabe2023tree}.
The naive constrained variants (GP-c, TPE-c) are informed of the feasible space by training them offline on 200 randomly sampled infeasible inputs with a penalty value assigned.
In these naive methods,
if an infeasible point is selected, it is assigned the worst possible evaluation value for each task.
The ablation study regarding the number of infeasible points sampled offline is described in Appendix~\ref{app:ablation_baselines}.
Also, as a conventional method for constrained tensor-based BBO, PROTES-c is included in the comparison~\cite{batsheva2023protes}.
Furthermore, we include an advanced method (NN+MILP-c)~\cite{papalexopoulos2022constrained}, which uses a piecewise-linear neural network as a surrogate and handles constraints via mixed-integer linear programming (MILP). 
For a comprehensive comparison, we also include the comparison methods run without task-specific constraints (NN+MILP-u / PROTES-u).

\subsection{Experiment 1:  HSDP vs. PGRAD}
\label{ssec:exp1}

\begin{figure}[h]
    \centering
    \includegraphics[width=\linewidth]{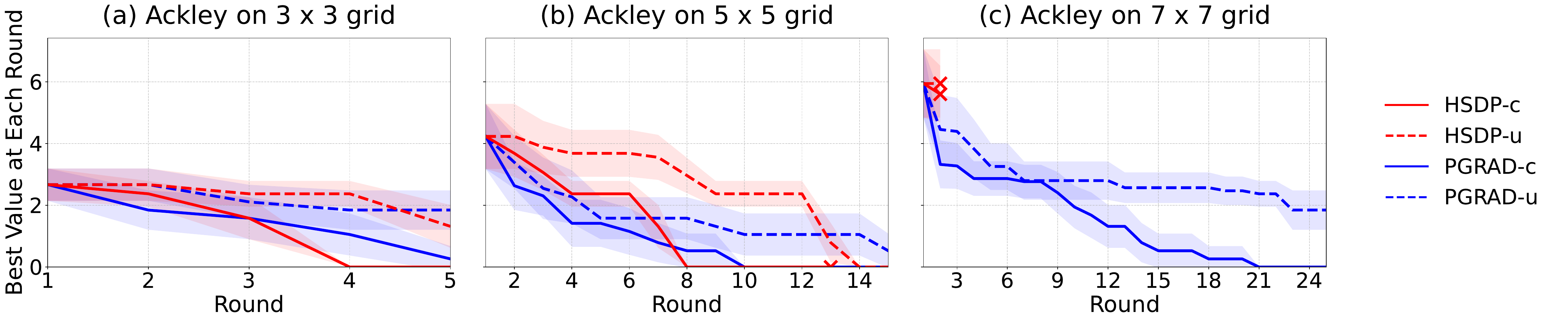}
    \vspace{-18pt}
    \caption{
        Optimization performance of CA-TD surrogates trained with HSDP and PGRAD on discrete Ackley benchmarks.  
        Lower and earlier curves indicate better sample efficiency, and narrower shaded areas reflect more stable performance across runs.
        Solid lines denote models trained with explicit feasibility integration, while dashed lines show unconstrained variants.  
        PGRAD (blue) offers better scalability and lower computational overhead, while HSDP (red) yields more stable convergence on small grids. 
    }
    \label{fig:exp1_conv}
    \vspace{-10pt}
\end{figure}

\begin{table}[ht]
    \centering
    \small 
    \setlength{\tabcolsep}{4pt}
    \caption{
        Comparison of HSDP and PGRAD for constraint-aware surrogate training on discrete Ackley benchmarks. 
        Each cell reports the mean over 10 runs, displaying the \textbf{Best Value} with the \textbf{Round} at  which it first appeared in subscripts using the notation $\textbf{Value}_{(\textbf{Round})}$. 
        Runtime indicates the average time in seconds per optimization round. 
        Standard deviations are omitted for brevity.
    }\label{tab:performance_by_constraint_model}
    \scriptsize
    \begin{tabular}{ll rr rr}
        \toprule
        & & \multicolumn{2}{c}{\textbf{Constrained (-c)}} & \multicolumn{2}{c}{\textbf{Unconstrained (-u)}} \\
        \cmidrule(lr){3-4} \cmidrule(lr){5-6}
        \textbf{Task} & \textbf{Method} & \textbf{Best Value$_{(\text{Round})}$} & \textbf{Runtime (s)} & \textbf{Best Value$_{(\text{Round})}$} & \textbf{Runtime (s)} \\
        \midrule
        Ackley & HSDP  & \textbf{\res{0.00}{3.40}} & 11.26 & \res{1.32}{3.00} & 11.49 \\
        $3\times3$ & PGRAD & \res{0.00}{3.70} & 1.76 & \textbf{\res{0.00}{6.70}} & 2.03 \\
        \midrule
        Ackley & HSDP  & \res{0.00}{6.90} & 279.44 & \res{0.00}{12.10} & 229.56 \\
        $5\times5$ & PGRAD & \textbf{\res{0.00}{5.60}} & 0.56 & \textbf{\res{0.00}{9.50}} & 1.44 \\
        \midrule
        Ackley & HSDP  & \res{5.60}{1.30} & 2045.80 & \res{5.95}{1.00} & 1932.85 \\
        $7\times7$ & PGRAD & \textbf{\res{0.00}{12.60}} & 0.67 & \textbf{\res{0.00}{25.40}} & 1.10 \\
        \bottomrule
    \end{tabular}
\end{table}

We compare two training methods for CA-TD: HSDP and PGRAD.
Both use TT format with rank $R=2$ and are tested on three Ackley grids.
For each method, we also include unconstrained counterparts trained without constraint-awareness.

Figure~\ref{fig:exp1_conv} shows the optimization progress.
On small grids, both constrained methods rapidly reach near-optimal values, with HSDP converging slightly earlier.
On the $7\times7$ grid, HSDP becomes impractical due to computational cost, while PGRAD continues to improve efficiently.
Table~\ref{tab:performance_by_constraint_model} summarizes best values, convergence rounds, and runtimes.
PGRAD achieves strong performance across all cases with runtimes under one second per round.
In contrast, HSDP is timed out (4000 seconds per round) in the case $7\times7$.
Across all cases, constraint versions  (``-c'') outperform unconstrained ones (``-u'') in both speed and final objective value.

\subsection{Experiment 2: Comparison using Benchmarks}
\label{ssec:exp2}

\begin{figure}[t]
    \centering
    \includegraphics[width=\linewidth]{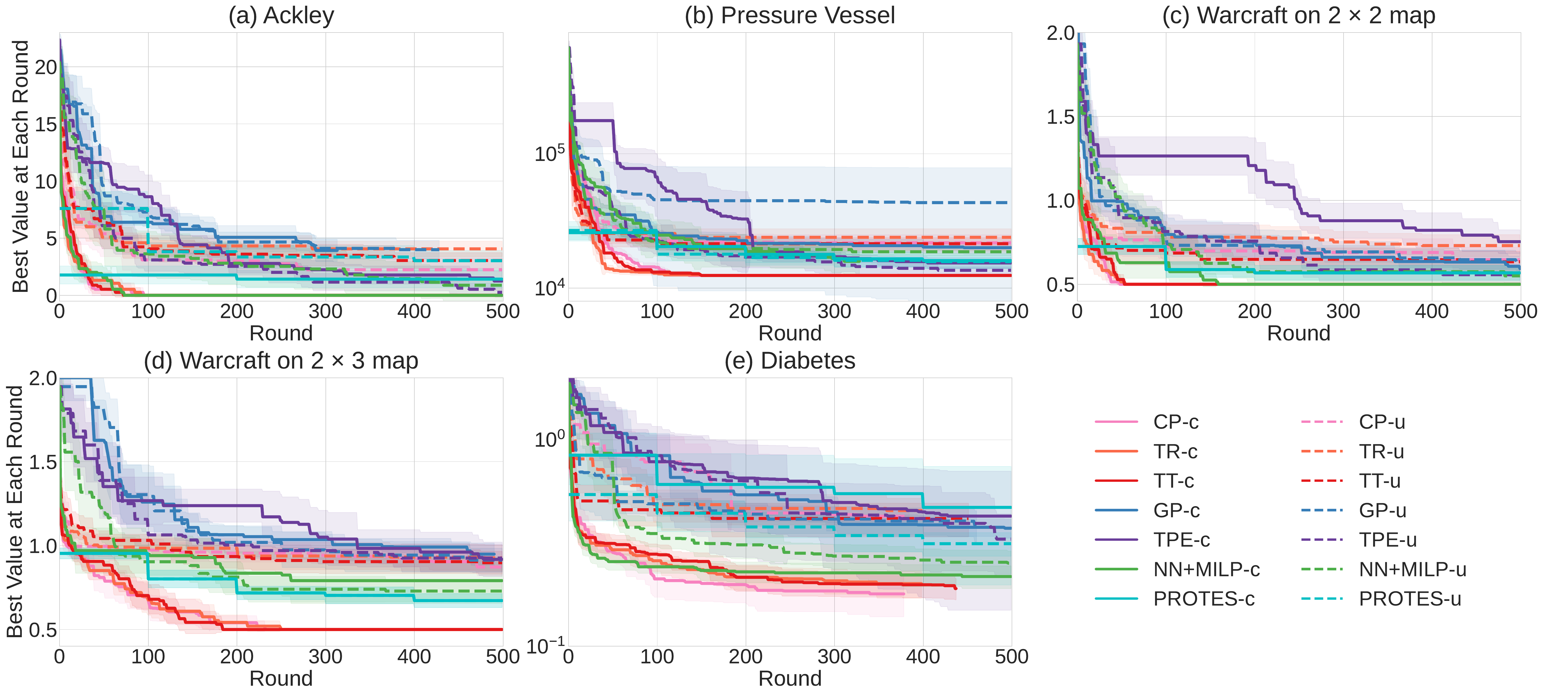}
    \vspace{-18pt}
    \caption{
        Optimization progress of our proposed tensor decomposition (TD) models and several baselines (GP, TPE, and NN+MILP), with and without constraint-awareness, across five benchmark tasks.
        For each TD model, the rank achieving the best performance is used (see Table~\ref{tab:main_summary} and Table~\ref{tab:baseline_summary} in the Appendix~\ref{app:reproducibility} for detailed results). 
        Lower and earlier curves indicate better sample efficiency.
        Solid lines denote constrained models (-c).
    }
    \label{fig:exp2_conv}
    \vspace{-10pt}
\end{figure}

In these experiments, we compare CA-TD with other methods using benchmarks.
Also, to examine how different TD formats affect the performance of CA-TD, we evaluate it under three formats: TT, CP and TR, each tested at ranks \(R = 2, \dots, 6\)~\ref{ssec:td}.
Throughout, constrained models are denoted with the suffix ``-c'', and unconstrained ones with ``-u'' where necessary.
For TT and TR, the same rank $R$ is uniformly applied across all modes. 
Figure~\ref{fig:exp2_conv} shows optimization curves for the best configuration of each method.
Our proposed CA-TD models, for example, TT-c, consistently achieve faster convergence and better final values compared to the naive baselines (GP-u, TPE-u, GP-c, and TPE-c). 
Crucially, CA-TD also demonstrates highly competitive or superior performance against the previous methods (PROTES and NN+MILP). 
This advantage is particularly evident in the Pressure Vessel and Warcraft benchmarks.
Note that NN+MILP is affected by hyperparameters such as the number of training epochs, and here we use the best hyperparameters (Appendix ~\ref{app:ablation_nn_milp}).

\subsection{Experiment 3: Extensive Comparison and Effect of Tensor Formats}

\begin{table}[t]
    \centering
    \caption{
        Performance comparison on benchmarks adopted from \cite{papalexopoulos2022constrained}. 
        For each tensor format, the rank $R \in \{2, \dots, 6\}$ that achieved the best objective value is reported. 
        Values are shown as $\textbf{Value}_{(\textbf{Round})}$. 
        The best-performing method for each task is highlighted in \textbf{bold}.
        (The numbers in parentheses outside the subscript represent the best rank R.)
    }\label{tab:main_comparison_compact_r}
    \footnotesize
    \setlength{\tabcolsep}{0.5pt} 
    \begin{tabular}{l rrr r}
        \toprule
        & \multicolumn{3}{c}{\textbf{CA-TD (Proposed)}} & \\
        \cmidrule(lr){2-4}
        \textbf{Task} & \textbf{CP (Best $R$)} & \textbf{TR (Best $R$)} & \textbf{TT (Best $R$)} & \textbf{NN+MILP-c} \\
        \midrule

        GAP A & \textbf{\res{-6.21}{99.1}} (5) & \res{-6.21}{149.0} (6) & \res{-6.21}{160.5} (6) & \res{-6.08}{243.0} \\
        GAP B & \textbf{\res{-4.19}{89.7}} (4) & \res{-4.19}{96.7} (4) & \res{-4.19}{101.8} (4) & \res{-4.12}{190.9} \\
        \midrule

        Ising A & \textbf{\res{-7.32}{76.0}} (6) & \res{-7.32}{120.2} (5) & \res{-7.32}{110.8} (5) & \res{-7.32}{319.0} \\
        Ising B & \textbf{\res{-9.43}{108.3}} (5) & \res{-9.43}{178.7} (5) & \res{-9.43}{201.6} (6) & \res{-8.92}{315.8} \\
        \midrule

        SSS & \res{-91.63}{176.2} (6) & \res{-91.66}{121.7} (5) & \res{-91.72}{104.9} (4) & \textbf{\res{-91.76}{208.4}} \\
        TSS & \res{-93.84}{270.0} (6) & \res{-93.84}{218.0} (2) & \textbf{\res{-93.84}{206.9}} (2) & \res{-93.75}{241.0} \\
        \midrule

        TfBind & \textbf{\res{-1.00}{270.4}} (3) & \res{-0.99}{376.8} (5) & \res{-0.99}{362.5} (5) & \res{-0.99}{297.5} \\
        \bottomrule
    \end{tabular}
\end{table}

To further verify the effectiveness of our framework, we conduct an extensive comparison against the strong NN+MILP baseline on the seven additional complex tasks. 
Moreover, to examine how different TD formats affect the performance of CA-TD, we evaluate CP, TR, and TT formats at varying ranks ($R\in\{2,\dots,6\}$).

Table~\ref{tab:main_comparison_compact_r} summarizes the best final objective values achieved by each method. 
The results demonstrate that our constraint-aware approach is highly competitive with the strong NN+MILP baseline across various combinatorial domains. 
More importantly, the evaluation of different tensor formats reveals distinct characteristics:
While the TT format served as a robust default, our results indicate that no single tensor format universally dominates all problem landscapes. 
The CP format identified the lowest objective values in specific tasks like GAP and Ising. 
Similarly, the TR format demonstrated task-dependent effectiveness, achieving performance levels comparable to the other evaluated formats and baselines in several instances. 
These findings underscore the flexibility of the CA-TD framework across different tensor representations.

\subsection{Discussion}
\label{ssec:exp_discussion}

The experiments validate the effectiveness of CA-TD for constrained black-box optimization. 
From Experiment~1, we confirm that incorporating feasibility into the training step of the surrogate model improves sample efficiency.
While HSDP performs well on small problems, PGRAD offers a scalable alternative suitable for larger settings such as the $7\times7$ Ackley grid.

From Experiment 2, we observe that our constraint-aware surrogate modeling is a key contributor in improving performance. 
Our CA-TD model consistently outperforms unconstrained optimization methods such as GP-u and TPE-u, and methods that simply include prior information, such as GP-c and TPE-c, and performs comparably to more advanced NN+MILP(-c) methods.
Among these, NN+MILP (both -c and -u) and GP-u/TPE-u are methods in which the surrogate model is trained without considering constraints, and constraints are considered only in the acquisition function stage.
Our results suggest this may lead to less efficient exploration.
By training the surrogate model to learn the boundaries of the feasible space, CA-TD can more accurately predict promising regions and improve sample efficiency.
This suggests that embedding feasibility directly into the surrogate model may be more effective than handling feasibility separately during acquisition.

Regarding the choice of tensor formats, our results highlight that there is no single ``best'' format for all landscapes. 
We adopted TT as a default due to its widespread use in existing tensor BBO literature, and it demonstrated robust and competitive performance across many tasks. 
However, as shown in Experiment 3, formats such as CP can outperform strong baselines on specific problems (e.g., GAP and Ising) when the optimal rank is identified.
Ultimately, these findings underscore that our proposed method to integrate logical constraints, which is to penalize constraint violations during surrogate training, is fundamentally effective regardless of the chosen tensor format providing a flexible and powerful framework for discrete constrained BBO.

A key limitation of CA-TD lies in its further scalability to high-dimensional search spaces.
While our experiments show speedups on PGRAD, further improvements in memory scalability are necessary to apply it to a wider range of problems.
Tensor-based BBO, including CA-TD, is limited by the memory demands of dense tensor representations. 
Although a simple mini-batching strategy offers a preliminary workaround (Appendix~\ref{app:minibatch}), fully scaling CA-TD to larger and higher-dimensional problems by using sparse tensor representation remains an important avenue for future research.

\section{Conclusion}
\label{sec:conclusion}

We proposed CA-TD, a constraint-aware surrogate modeling approach for sequential black-box optimization on discrete domains, which integrates symbolic feasibility information directly into tensor decomposition-based surrogate models.
By formulating the learning problem as a POP and introducing a relaxed gradient-based algorithm (PGRAD), we enabled scalable neuro-symbolic integration.
Extensive experiments across diverse synthetic and real-world benchmarks demonstrated that CA-TD improves sample efficiency compared to conventional methods. 
Crucially, by embedding constraints into the surrogate rather than relying solely on acquisition optimization, our approach achieved performance comparable or superior to strong baselines such as NN+MILP. 
Our evaluation also highlighted the flexibility of the CA-TD framework regarding tensor formats: while the TT format serves as a robust and stable default, formats like CP can achieve the highest optimization performance on specific problem landscapes when the rank is appropriately tuned.

Future work includes scaling to higher-dimensional discrete spaces, for instance, via sparse tensor representations.
While the TT format already mitigates scalability issues, automatic rank selection would further enhance the applicability of these tensor formats. 
Extending this logic-integrated approach to continuous domains and mixed-variable BBO problems remains an exciting avenue for broadening its real-world impact.

\bibliographystyle{splncs04}
\bibliography{references}

\clearpage

\appendix
\renewcommand{\theHsection}{appendix.\Alph{section}}

\section{Specifications of Benchmark Problems}
\label{app:benchmarks}

\begin{table}[htbp]
  \caption{Search spaces for each task}
  \label{tab:benchmark_search_space}
  \centering
  \begin{tabular}{lccc}
    \toprule
    task &search space size & \#feasible points &ratio of feasible points \\
    \hline
    Ackley 3$\times$3   & $3^2$     & 5     & 0.56 \\
    Ackley 5$\times$5   & $5^2$     & 13    & 0.52 \\
    Ackley 7$\times$7   & $7^2$     & 29    & 0.59 \\ 
    \hline
    Ackley 65$\times$65 & $65^2$    & 317   & 0.08 \\
    Pressure Vessel     & $10^4$    & 3916  & 0.39 \\
    Warcraft 2$\times$2 & $7^4$     & 300   & 0.12 \\
    Warcraft 2$\times$3 & $7^6$     & 5400  & 0.05 \\
    Diabetes     & $5^8$     & 10197 & 0.03 \\
    \midrule
    GAP A               & $3^9$     & 1260  & 0.06 \\
    GAP B               & $4^7$     & 2368  & 0.14 \\
    Ising A             & $2^{14}$  & 441   & 0.03 \\
    Ising B             & $2^{15}$  & 1260  & 0.04 \\
    NAS TSS             & $5^6$     & 1520  & 0.10 \\
    NAS SSS             & $8^5$     & 6552  & 0.20 \\
    TfBind              & $4^8$     & 23808 & 0.36 \\
    \bottomrule
  \end{tabular}
\end{table}

This section provides details of the benchmark tasks used in our experiments.
The differences in each search space are shown in Table~\ref{tab:benchmark_search_space}.
Details of each task are described below.

\subsection{Benchmarks for Experiment 2}

\paragraph{Ackley}
The 2D Ackley function~\cite{adorio2005ackley} on grid is defined as:
\begin{align*}
g(x_1, x_2) &= -20 e^{-0.2\sqrt{0.5(x_1^2 + x_2^2)}}  - e^{0.5(\cos(2\pi x_1) + \cos(2\pi x_2))} + 20 + e.
\end{align*}
The input space is discretized into a uniform integer grid, and feasibility is defined by a circular constraint \(x_1^2 + x_2^2 \leq r^2\).

\paragraph{Pressure Vessel}
The Pressure Vessel design problem is a classic engineering benchmark~\cite{coello2002constraint} with a mixed-variable search space. 
The goal is to minimize the total cost of a cylindrical pressure vessel. 
The problem has four variables, originally two continuous and two integer. 
For our experiments, we create a fully discrete search space by sampling 10 uniform levels from the domain of each variable. The objective function is given by:
\begin{align*}
    g(x_1, x_2, x_3, x_4) &= 0.6224 x_1 x_3 x_4 + 1.7781 x_2 x_3^2 + 3.1661 x_1^2 x_4 + 19.84 x_1^2 x_3,
\end{align*}
subject to the following inequality constraints:
$-x_1 + 0.0193 x_3 \leq 0$, 
$-x_2 + 0.00954 x_3 \leq 0$, 
$-\pi x_3^2 x_4 - \frac{4}{3}\pi x_3^3 + 1296000 \leq 0$, 
and $x_4 - 240 \leq 0$.

\paragraph{Warcraft}
This benchmark, adapted from a path prediction problem solved via supervised learning with combinatorial constraints presented in~\cite{ahmed2022warcraft}, is treated as a black-box optimization task.
The environment is a 2D \( m \times n \) grid map, where each cell has a predefined traversal cost.  
An input \(\mathbf{x} \in \mathcal{X}\), representing a candidate path, is encoded as a sequence of \(m + n\) movement primitives.  
Each movement primitive is selected from a set of seven types $\mathcal{A} = \{a_1, \dots, a_7\}$, which define the displacement in the 2D grid: cardinal moves (up, down, left, right), two diagonal-like L-shaped moves (e.g., one step up followed by one step right), and a null move representing no displacement. 
Formally, these primitives correspond to the set of vectors $\{(0, 1), (0, -1), (-1, 0), (1, 0), (1, 1), (-1, -1), (0, 0)\}$.

The objective function \(g(\mathbf{x})\) evaluates each path by summing the traversal costs along the path and rewarding proximity to the bottom-right corner, with shorter Euclidean distance yielding better scores.
The input is subject to three constraints: the path must start from the top-left cell, it must consist of exactly \(m + n\) steps, and it must end at the bottom-right cell.

We evaluate two map sizes: \(2 \times 2\) (path length 4) and \(2 \times 3\) (path length 6), resulting in \(7^4\) and \(7^6\) candidate paths, respectively.

\paragraph{Diabetes} 
This task simulates the goal of identifying actionable treatment plans for patients diagnosed with diabetes to demonstrate applicability on a real-world task where constraints are derived from domain knowledge.
We use the Pima Indian Diabetes dataset~\cite{smith1988using}, which contains 8 patient features and a binary diabetes label.  
Each continuous or integer-valued feature is discretized into 5 levels, resulting in a discrete search space \(\mathcal{X} = \{0,1,2,3,4\}^8\).  

A random forest classifier~\cite{breiman2001random} is trained on the entire dataset and used to predict the probability of diabetes for all candidates \(\mathbf{x} \in \mathcal{X}\).  
Given a randomly chosen diabetic individual \(\mathbf{x}_{\text{orig}}\), the goal is to find an alternative feature configuration \(\mathbf{x} \in \mathcal{X}\) such that the predicted probability of diabetes is reduced.  
To encourage realistic plans, we penalize large deviations from the original configuration.  
Specifically, the objective function is defined as:
\[
g(\mathbf{x}) = \text{RF}(\mathbf{x}) + \|\mathbf{x} - \mathbf{x}_{\text{orig}}\|_2,
\]
where \(\text{RF}(\mathbf{x}) \in [0,1]\) is the predicted probability from the random forest classifier.  
Lower objective values correspond to medically plausible and effective treatment suggestions.

Feasibility constraints \(c(\mathbf{x}) = 1\) are imposed to exclude unrealistic feature combinations by enforcing logical consistency based on domain heuristics. 
For example, we define the feasible region to exclude physiologically contradictory states, such as \(\neg (\text{Insulin} < \theta_1 \land \text{Glucose} > \theta_2)\), where \(\theta_1\) and \(\theta_2\) are thresholds for clinical implausibility. 
This setup evaluates the ability of CA-TD to internalize symbolic prior knowledge in a real-world-inspired search space.

\subsection{Benchmarks for Experiment 3}

We tailored seven tasks across four classes evaluated in~\cite{papalexopoulos2022constrained} by adding constraints and down-scaling the tasks to evaluate CA-TD against NN+MILP. 
All maximize a given metric, formulated here as minimizing its negative.

\paragraph{Generalized Assignment Problem (GAP)} 
Assign $d$ items to $m$ bins to maximize total value $Value(\mathbf{x})=\sum_{i=1}^{d}p_{i,x_{i}}$, where $x_i \in \{1, \dots, m\}$ is the bin assigned to item $i$, and $p_{i,j}$ is the profit of assigning item $i$ to bin $j$. $\mathbb{I}(\cdot)$ denotes the indicator function.
\begin{itemize}
    \item \textbf{GAP-A} ($3^9$): Capacity-constrained $\sum_{i=1}^{d} w_{i} \cdot \mathbb{I}(x_{i}=j)=c_{j}$ for each bin $j$, where $w_i=1$ is the weight of item $i$, and $c_j$ is the capacity of bin $j$.
    \item \textbf{GAP-B} ($4^7$): Logically-constrained $\sum_{i=1}^{d} \mathbb{I}(x_{i}=1)\le1$ and $\sum_{i=1}^{d} \mathbb{I}(x_{i}=2)\le1$.
\end{itemize}

\paragraph{Constrained Ising Model} 
Minimize $f(\mathbf{y})=\sum_{i=1}^{d-1}\sum_{j=i+1}^{d}P_{ij}y_{i}y_{j}$ over binary vectors $\mathbf{y} \in \{0, 1\}^d$, where $y_i=1$ indicates that item $i$ is selected, and $P_{ij}$ is the interaction potential between items $i$ and $j$.
\begin{itemize}
    \item \textbf{Ising-A} ($2^{14}$): Group balance constraints where item groups must have equal selection counts and a total cardinality of $\sum_{i=1}^{d} y_i = 4$.
    \item \textbf{Ising-B} ($2^{15}$): Complex constraints where group A selection equals group B, and group C must have exactly 1 item.
\end{itemize}

\paragraph{Neural Architecture Search (NAS)} 
Maximize test accuracy on CIFAR-10 from NATS-Bench.
\begin{itemize}
    \item \textbf{TSS} ($5^6$): Selecting operations $z_i$ for $6$ edges with constraints $\sum_{i=1}^{6} \mathbb{I}(z_i = \text{`skip\_connect'}) \ge 3$ and $\sum_{i=1}^{6} \mathbb{I}(z_i = \text{`nor\_conv\_3x3'}) \le 2$.
    \item \textbf{SSS} ($8^5$): Selecting channel counts $z_i$ for $5$ stages constrained by $\sum_{i=1}^{5} z_i \le 160$ and $z_4 \ge z_2$.
\end{itemize}

\paragraph{DNA Binding (TfBind)} 
Optimizing binding affinity for a DNA sequence $\mathbf{x} = (x_1, \dots, x_8)$ of length $8$ ($4^8$ search space), where $x_i \in \{\text{A}, \text{C}, \text{G}, \text{T}\}$. The GC-content is restricted by the cardinality constraint: $\sum_{i=1}^{8} \mathbb{I}(x_{i} \in \{\text{G}, \text{C}\}) \le 3$.

\section{Experimental Details}
\label{app:reproducibility}

All experiments were conducted on nodes running Ubuntu 22.04.5 LTS. Each experimental run was allocated 4 cores of an Intel Xeon Gold 6230R CPU and 8 GB of memory. A timeout of 3600 seconds (1 hour) was set for each run.

The software environment was built on Python 3.12.2. Key libraries include PyTorch 2.4.1 and NumPy 2.1.2. Our HSDP training strategy utilized ncpol2sdpa 1.12.2 and cvxpy 1.6.4, with SDPA 7.3.16 as the backend semidefinite programming solver. Our implementation of the NN+MILP baseline~\cite{papalexopoulos2022constrained} follows the methodology described in the original paper. The mixed-integer linear programming subproblems are solved using OR-Tools 9.14.6206.

\section{Detailed Experimental Results}
\label{app:full_results}

Here, we provide the complete summary tables compiling the results of our experiments. 
Table~\ref{tab:main_summary} and Table~\ref{tab:baseline_summary} show the full performance details of our proposed methods and the baselines for the initial five benchmark tasks. 
Table~\ref{tab:nnmilp_full_comparison_all_ranks} presents the comprehensive results across all evaluated tensor ranks for the advanced comparison tasks against the NN+MILP baseline.

\begin{sidewaystable}[p]
    \centering
    \small 
    \setlength{\tabcolsep}{3pt}
    \caption{
        Full experimental results comparing CP, TR, and TT formats across different ranks. 
        Each cell reports the mean over 10 runs, displaying the \textbf{Best Value} with the \textbf{Round} at  which it first appeared in subscripts using the notation $\textbf{Value}_{(\textbf{Round})}$. 
        Bold indicates the best rank within each method. 
        Standard deviations are omitted for brevity.
    }\label{tab:main_summary}
    \begin{tabular}{ll ccc ccc}
        \toprule
        & & \multicolumn{3}{c}{\textbf{Constrained (-c)}} & \multicolumn{3}{c}{\textbf{Unconstrained (-u)}} \\
        \cmidrule(lr){3-5} \cmidrule(lr){6-8}
        \textbf{Task} & \textbf{R} & \textbf{CP} & \textbf{TR} & \textbf{TT} & \textbf{CP} & \textbf{TR} & \textbf{TT} \\
        \midrule

        Ackley & 2 & \res{0.00}{69.40} & \res{0.00}{55.80} & \res{0.00}{58.00} & \res{5.00}{74.00} & \textbf{\res{4.07}{68.50}} & \res{4.74}{151.60} \\
               & 3 & \textbf{\res{0.00}{37.70}} & \res{0.00}{61.60} & \textbf{\res{0.00}{36.30}} & \res{4.37}{179.50} & \res{5.72}{104.90} & \res{3.88}{135.70} \\
               & 4 & \res{0.00}{50.20} & \textbf{\res{0.00}{48.60}} & \res{0.00}{47.50} & \res{5.11}{61.30} & \res{5.31}{47.80} & \res{3.70}{113.20} \\
               & 5 & \res{0.00}{56.60} & \res{0.00}{57.70} & \res{0.00}{57.40} & \res{4.29}{76.80} & \res{5.01}{79.00} & \res{4.38}{111.30} \\
               & 6 & \res{0.00}{63.50} & \res{0.00}{71.90} & \res{0.00}{63.70} & \textbf{\res{2.24}{105.10}} & \res{6.05}{27.30} & \textbf{\res{3.05}{138.40}} \\
        \midrule

        Diabetes & 2 & \res{0.26}{190.30} & \res{0.26}{193.50} & \res{0.27}{164.60} & \res{0.56}{237.60} & \textbf{\res{0.45}{82.60}} & \res{0.59}{240.50} \\
                 & 3 & \res{0.20}{245.40} & \res{0.25}{205.30} & \res{0.25}{215.40} & \res{0.44}{105.70} & \res{0.49}{142.30} & \textbf{\res{0.42}{36.70}} \\
                 & 4 & \res{0.22}{215.10} & \res{0.22}{276.70} & \res{0.19}{289.50} & \res{0.49}{26.60} & \res{0.49}{21.80} & \res{0.43}{79.30} \\
                 & 5 & \res{0.19}{229.20} & \res{0.20}{263.60} & \res{0.21}{242.30} & \textbf{\res{0.43}{139.20}} & \res{0.59}{33.60} & \res{0.51}{24.30} \\
                 & 6 & \textbf{\res{0.18}{183.40}} & \textbf{\res{0.20}{210.00}} & \textbf{\res{0.19}{258.00}} & \res{0.44}{163.30} & \res{0.50}{62.80} & \res{0.43}{76.40} \\
        \midrule

        Pressure & 2 & \textbf{\res{12408.34}{97.60}} & \textbf{\res{12408.34}{65.60}} & \res{12408.34}{98.20} & \res{28747.63}{135.10} & \res{28563.29}{42.20} & \res{26976.59}{36.70} \\
        Vessel   & 3 & \res{12408.34}{101.60} & \res{12408.34}{107.90} & \textbf{\res{12408.34}{91.80}} & \res{26880.85}{227.20} & \res{24871.78}{65.80} & \res{26170.20}{65.20} \\
                 & 4 & \res{12408.34}{114.10} & \res{12408.34}{153.90} & \res{12408.34}{120.90} & \res{27551.16}{166.20} & \res{32783.74}{36.00} & \res{22564.40}{67.70} \\
                 & 5 & \res{12408.34}{125.10} & \res{12408.34}{171.60} & \res{12408.34}{186.70} & \res{24153.88}{63.90} & \res{34158.80}{19.20} & \res{22935.84}{28.40} \\
                 & 6 & \res{12408.34}{133.60} & \res{12408.34}{158.60} & \res{12408.34}{162.70} & \textbf{\res{21718.03}{83.80}} & \textbf{\res{23866.31}{28.70}} & \textbf{\res{21395.48}{34.60}} \\
        \midrule

        Warcraft & 2 & \res{0.50}{40.80} & \res{0.50}{29.60} & \res{0.50}{34.90} & \res{0.67}{192.30} & \textbf{\res{0.73}{171.40}} & \res{0.65}{192.30} \\
        $2\times2$ & 3 & \res{0.50}{31.00} & \res{0.50}{29.70} & \textbf{\res{0.50}{34.10}} & \res{0.74}{108.90} & \res{0.74}{188.20} & \res{0.75}{213.00} \\
                 & 4 & \res{0.50}{31.30} & \res{0.50}{33.00} & \res{0.50}{39.80} & \res{0.68}{212.00} & \res{0.74}{163.90} & \res{0.71}{181.40} \\
                 & 5 & \res{0.50}{37.90} & \textbf{\res{0.50}{26.10}} & \res{0.50}{41.60} & \textbf{\res{0.65}{96.30}} & \res{0.76}{123.70} & \res{0.67}{195.80} \\
                 & 6 & \textbf{\res{0.50}{25.60}} & \res{0.50}{41.80} & \res{0.50}{47.20} & \res{0.68}{181.20} & \res{0.76}{195.30} & \textbf{\res{0.63}{115.60}} \\
        \midrule

        Warcraft & 2 & \res{0.56}{108.10} & \res{0.56}{335.20} & \res{0.58}{257.90} & \res{0.97}{221.70} & \textbf{\res{0.93}{130.20}} & \textbf{\res{0.90}{184.60}} \\
        $2\times3$ & 3 & \res{0.50}{145.20} & \textbf{\res{0.50}{139.10}} & \res{0.50}{116.90} & \textbf{\res{0.87}{234.50}} & \res{1.01}{92.40} & \res{0.95}{204.20} \\
                 & 4 & \textbf{\res{0.50}{123.00}} & \res{0.51}{255.50} & \textbf{\res{0.50}{115.70}} & \res{0.90}{96.90} & \res{1.04}{150.30} & \res{0.98}{167.80} \\
                 & 5 & \res{0.50}{125.70} & \res{0.50}{144.50} & \res{0.50}{144.70} & \res{0.88}{148.50} & \res{0.94}{120.90} & \res{1.01}{148.90} \\
                 & 6 & \res{0.50}{126.90} & \res{0.50}{175.20} & \res{0.50}{152.80} & \res{0.88}{156.10} & \res{0.93}{91.50} & \res{0.96}{56.90} \\
        \bottomrule
    \end{tabular}
\end{sidewaystable}

\begin{table}[htbp]
    \centering
    \small 
    \setlength{\tabcolsep}{6pt} %
    \caption{
        Performance summary of the baseline methods (GP, TPE, NN+MILP, and PROTES) in Experiment 2. 
        Each cell reports the mean over 10 runs, displaying the \textbf{Best Value} with the \textbf{Round} at  which it first appeared in subscripts using the notation $\textbf{Value}_{(\textbf{Round})}$. 
        Standard deviations are omitted for brevity as per the main CA-TD results.
    }\label{tab:baseline_summary}

    \begin{tabular}{ll rr}
        \toprule
        & & \multicolumn{1}{c}{\textbf{Constrained (-c)}} & \multicolumn{1}{c}{\textbf{Unconstrained (-u)}} \\
        \cmidrule(lr){3-3} \cmidrule(lr){4-4}
        \textbf{Task} & \textbf{Method} & \textbf{Best Value$_{(\text{Round})}$} & \textbf{Best Value$_{(\text{Round})}$} \\
        \midrule

        Ackley   & GP      & \res{3.72}{249.90} & \res{4.00}{209.40} \\
                 & TPE     & \res{1.25}{295.50} & \res{0.26}{254.10} \\
                 & NN+MILP & \res{0.00}{41.30}  & \res{0.88}{222.30} \\
                 & PROTES  & \res{1.42}{140.00} & \res{3.03}{280.00} \\
        \midrule

        Diabetes & GP      & \res{0.37}{262.20} & \res{0.39}{208.80} \\
                 & TPE     & \res{0.43}{312.30} & \res{0.33}{388.80} \\
                 & NN+MILP & \res{0.22}{176.60} & \res{0.25}{334.3} \\
                 & PROTES  & \res{0.47}{410.00} & \res{0.31}{380.00} \\
        \midrule

        Pressure & GP      & \res{19982.64}{250.30} & \res{43256.59}{169.70} \\
        Vessel   & TPE     & \res{15375.32}{294.10} & \res{13550.90}{238.20} \\
                 & NN+MILP & \res{14394.76}{239.20} & \res{18632.64}{161.10} \\
                 & PROTES  & \res{16006.94}{350.00} & \res{15704.22}{340.00} \\
        \midrule

        Warcraft & GP      & \res{0.59}{307.00} & \res{0.64}{271.50} \\
        2x2      & TPE     & \res{0.75}{288.30} & \res{0.56}{194.10} \\
                 & NN+MILP & \res{0.50}{66.10}  & \res{0.55}{150.60} \\
                 & PROTES  & \res{0.57}{190.00} & \res{0.57}{190.00} \\
        \midrule

        Warcraft & GP      & \res{0.95}{225.60} & \res{0.91}{231.30} \\
        2x3      & TPE     & \res{0.92}{280.10} & \res{0.93}{169.50} \\
                 & NN+MILP & \res{0.79}{85.00}  & \res{0.73}{151.40} \\
                 & PROTES  & \res{0.67}{360.00} & \res{0.67}{360.00} \\
        \bottomrule
    \end{tabular}
\end{table}

\begin{sidewaystable}[p]
    \centering
    \setlength{\tabcolsep}{3pt}
    \caption{
        Full experimental results on benchmarks from \cite{papalexopoulos2022constrained} across all evaluated ranks $R \in \{2, \dots, 6\}$. 
        Values are reported as $\textbf{Value}_{(\textbf{Round})}$. 
        The best performing configuration for each task (lowest value, then earliest round) is highlighted in \textbf{bold}.
    }\label{tab:nnmilp_full_comparison_all_ranks}
    \begin{tabular}{ll ccccc r}
        \toprule
        \textbf{Task} & \textbf{Format} & \textbf{R=2} & \textbf{R=3} & \textbf{R=4} & \textbf{R=5} & \textbf{R=6} & \textbf{NN+MILP-c} \\
        \midrule

        \multirow{3}{*}{GAP A} 
        & CP    & \res{-6.17}{220.1} & \res{-6.21}{283.3} & \res{-6.21}{124.3} & \textbf{\res{-6.21}{99.1}} & \res{-6.21}{102.6} & \multirow{3}{*}{\res{-6.08}{243.0}} \\
        & TR    & \res{-6.07}{131.1} & \res{-6.03}{211.9} & \res{-6.17}{282.8} & \res{-6.21}{229.7} & \res{-6.21}{149.0} & \\
        & TT    & \res{-5.90}{179.7} & \res{-5.97}{150.5} & \res{-6.16}{292.3} & \res{-6.21}{192.5} & \res{-6.21}{160.5} & \\
        \midrule

        \multirow{3}{*}{GAP B} 
        & CP    & \res{-4.15}{258.2} & \res{-4.19}{120.9} & \textbf{\res{-4.19}{89.7}} & \res{-4.19}{113.2} & \res{-4.19}{105.4} & \multirow{3}{*}{\res{-4.12}{190.9}} \\
        & TR    & \res{-4.17}{182.1} & \res{-4.19}{167.2} & \res{-4.19}{96.7} & \res{-4.19}{116.6} & \res{-4.19}{118.0} & \\
        & TT    & \res{-4.18}{224.0} & \res{-4.19}{134.5} & \res{-4.19}{101.8} & \res{-4.19}{123.4} & \res{-4.19}{110.5} & \\
        \midrule

        \multirow{3}{*}{Ising A} 
        & CP    & \res{-7.27}{127.6} & \res{-7.32}{173.6} & \res{-7.32}{120.2} & \res{-7.32}{85.1} & \textbf{\res{-7.32}{76.0}} & \multirow{3}{*}{\res{-7.32}{319.0}} \\
        & TR    & \res{-7.27}{184.5} & \res{-7.32}{152.9} & \res{-7.32}{156.5} & \res{-7.32}{120.2} & \res{-7.32}{181.4} & \\
        & TT    & \res{-7.32}{124.2} & \res{-7.32}{171.2} & \res{-7.32}{111.6} & \res{-7.32}{110.8} & \res{-7.32}{138.4} & \\
        \midrule

        \multirow{3}{*}{Ising B} 
        & CP    & \res{-9.32}{202.0} & \res{-9.43}{258.1} & \res{-9.43}{141.2} & \textbf{\res{-9.43}{108.3}} & \res{-9.43}{118.5} & \multirow{3}{*}{\res{-8.92}{315.8}} \\
        & TR    & \res{-9.23}{175.9} & \res{-9.34}{394.3} & \res{-9.43}{258.7} & \res{-9.43}{178.7} & \res{-9.43}{205.5} & \\
        & TT    & \res{-9.31}{119.7} & \res{-9.06}{357.6} & \res{-9.43}{232.9} & \res{-9.43}{206.8} & \res{-9.43}{201.6} & \\
        \midrule

        \multirow{3}{*}{SSS} 
        & CP    & \res{-91.52}{132.1} & \res{-91.52}{66.3} & \res{-91.48}{102.1} & \res{-91.57}{199.9} & \res{-91.63}{176.2} & \multirow{3}{*}{\textbf{\res{-91.76}{208.4}}} \\
        & TR    & \res{-91.54}{43.0} & \res{-91.46}{47.4} & \res{-91.59}{133.1} & \res{-91.66}{121.7} & \res{-91.65}{202.0} & \\
        & TT    & \res{-91.40}{44.1} & \res{-91.55}{67.1} & \res{-91.72}{104.9} & \res{-91.61}{191.7} & \res{-91.57}{208.4} & \\
        \midrule

        \multirow{3}{*}{TSS} 
        & CP    & \res{-93.73}{136.1} & \res{-93.81}{303.2} & \res{-93.81}{205.6} & \res{-93.82}{203.1} & \res{-93.84}{270.0} & \multirow{3}{*}{\res{-93.75}{241.0}} \\
        & TR    & \res{-93.84}{218.0} & \res{-93.84}{227.6} & \res{-93.82}{255.9} & \res{-93.83}{294.6} & \res{-93.83}{273.8} & \\
        & TT    & \textbf{\res{-93.84}{206.9}} & \res{-93.84}{247.1} & \res{-93.83}{253.8} & \res{-93.84}{332.0} & \res{-93.83}{230.5} & \\
        \midrule

        \multirow{3}{*}{TfBind} 
        & CP    & \res{-0.99}{281.5} & \textbf{\res{-1.00}{270.4}} & \res{-0.99}{390.7} & \res{-1.00}{330.4} & \res{-1.00}{322.1} & \multirow{3}{*}{\res{-0.99}{297.5}} \\
        & TR    & \res{-0.98}{204.9} & \res{-0.98}{231.7} & \res{-0.99}{311.1} & \res{-0.99}{376.8} & \res{-0.99}{308.1} & \\
        & TT    & \res{-0.98}{303.8} & \res{-0.98}{270.4} & \res{-0.99}{341.8} & \res{-0.99}{362.5} & \res{-0.99}{357.8} & \\
        \bottomrule
    \end{tabular}
\end{sidewaystable}

\section{Ablation study for Experiment 2}
\label{app:addtional_results}

\subsection{Effect of the Penalty Coefficient $\lambda$}
\label{app:ablation_lambda}

\begin{figure}[ht]
    \centering
    \begin{minipage}{0.9\linewidth}
        \centering
        \includegraphics[width=\linewidth]{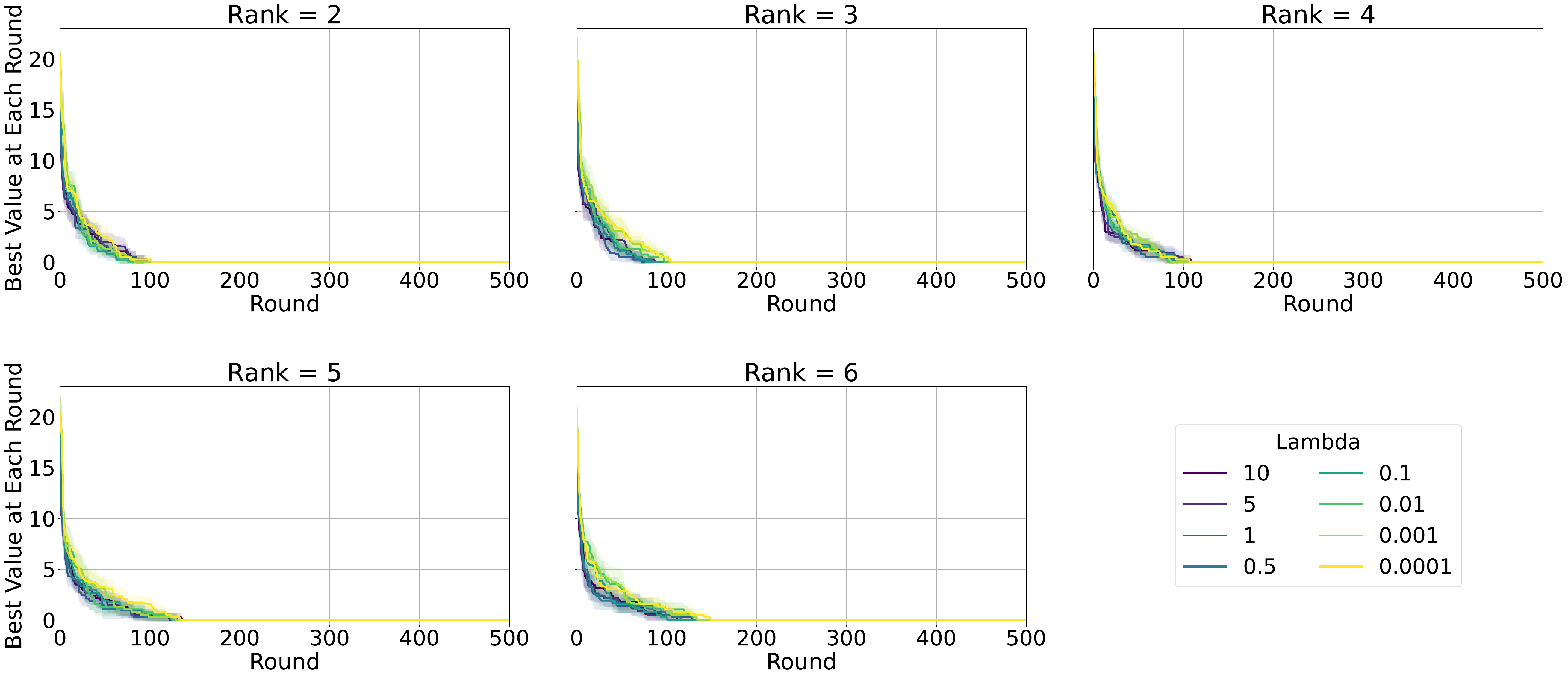} \\
        \small (a) Ackley on $65\times65$ grid
    \end{minipage}
    
    \vspace{15pt} %
    
    \begin{minipage}{0.9\linewidth}
        \centering
        \includegraphics[width=\linewidth]{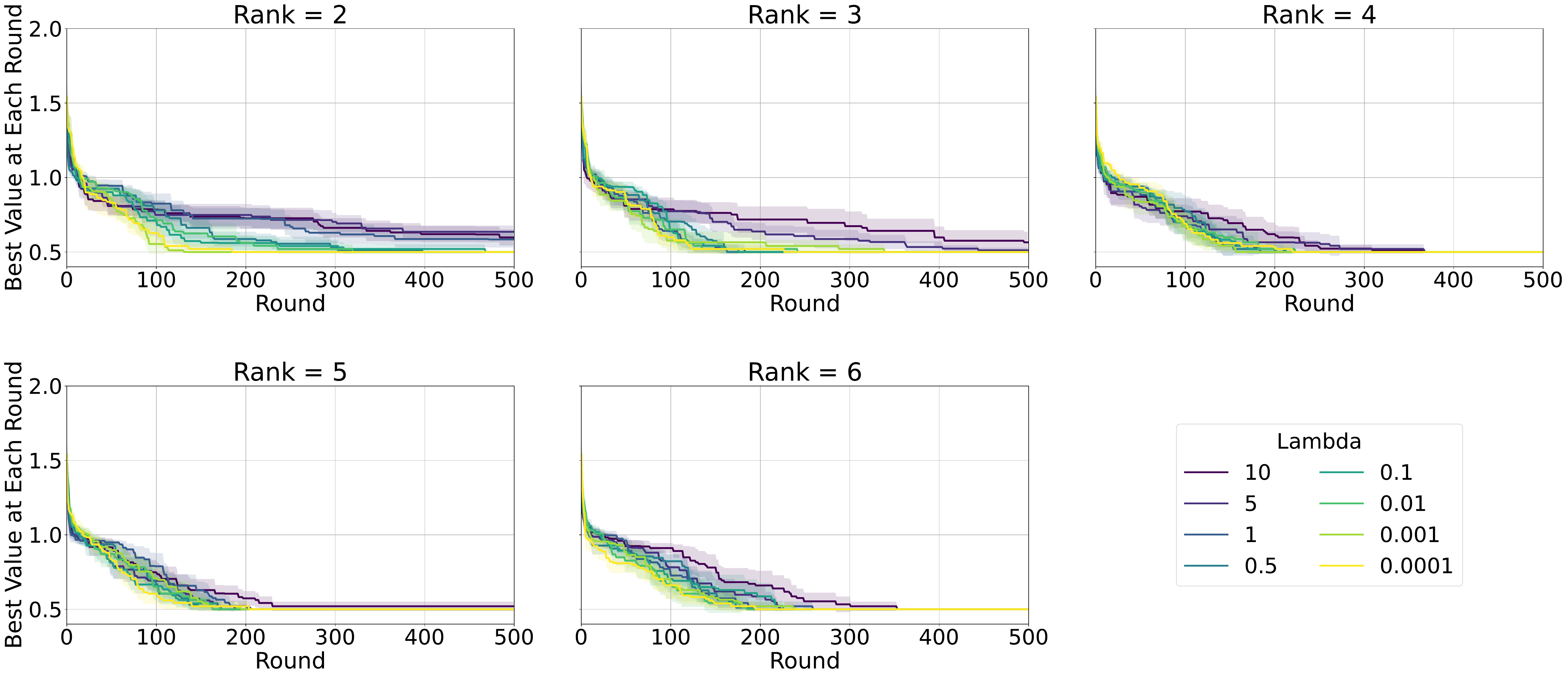} \\
        \small (b) Warcraft on $2\times3$ map
    \end{minipage}
    
    \caption{
        Representative sensitivity analysis of the penalty coefficient $\lambda$ for TT-c. 
        (a) Ackley illustrates the typical high robustness found in the majority of tasks. 
        (b) Warcraft $2\times3$ shows a slight performance sensitivity specifically at very high $\lambda$ values. 
        The overlapping curves justify the choice of $\lambda=1.0$ as a robust default.
    }
    \label{fig:ablation_lambda_rep}
\end{figure}

We evaluated the sensitivity of the PGRAD strategy to the penalty coefficient $\lambda$, varying it from $0.0001$ to $10$ for CP-c, TR-c, and TT-c surrogates across all benchmarks. 
As shown in Figure~\ref{fig:ablation_lambda_rep}, the performance is remarkably robust to this hyperparameter, with nearly identical convergence behavior across a wide range of values. 
This high stability justifies our choice of $\lambda=1.0$ as a standard setting for CA-TD.

A minor exception was observed in the Warcraft $2\times3$ map benchmark (Figure~\ref{fig:ablation_lambda_rep}b), where very high penalty coefficients ($\lambda=10$ and $\lambda=5$) resulted in slightly slower initial convergence. 
This is likely due to the increased difficulty of the landscape when constraints are overly penalized. 
However, even in this case, the method remains competitive, confirming that CA-TD does not require extensive tuning of $\lambda$ to achieve stable performance.

\subsection{Effect of Offline Training on GP and TPE Baselines}
\label{app:ablation_baselines}

In our main experiments, the constrained GP and TPE baselines (GP-c and TPE-c) were trained with 200 offline-sampled infeasible inputs. 
To analyze the impact of this prior knowledge, we varied the number of infeasible points in $\{0, 50, 100, 200, 300, 500, 1000, 2000\}$ across all benchmarks. 
We observed a consistent, counter-intuitive trend: increasing the number of pre-trained infeasible points generally degrades optimization performance.

\begin{figure}[ht]
    \centering
    \includegraphics[width=0.9\linewidth]{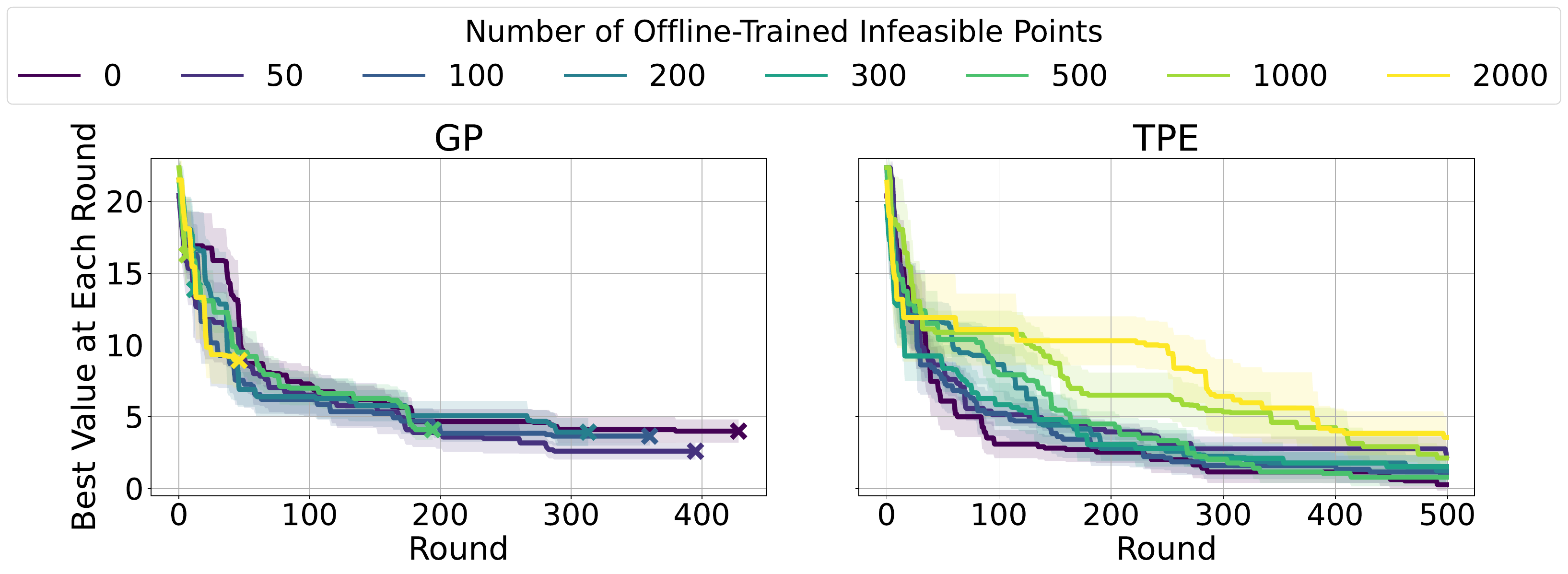}
    \caption{
        Representative ablation study (Ackley $65\times65$ grid) on the number of pre-trained infeasible points for GP-c and TPE-c. 
        Left and right panels show GP-c and TPE-c, respectively. 
        Increasing informed points consistently degraded performance across all five tested benchmarks.
    }
    \label{fig:ablation_gp_tpe_rep}
\end{figure}

While this trend was observed across all five tasks, Figure~\ref{fig:ablation_gp_tpe_rep} presents the Ackley $65 \times 65$ grid as a representative case. 
The degradation is particularly severe for the GP baseline, where computational overhead leads to truncated convergence curves and premature termination. 
This is due to the GP model's cubic scaling, which makes it prohibitively expensive for acquisition function optimization as data points increase. 
The TPE baseline proved more computationally robust but still exhibited similar performance decay. 
These findings suggest that naively expanding the training dataset with infeasible points is ineffective and potentially detrimental for baseline models.

\subsection{Effect of Number of Epochs and Initial Points for NN+MILP}
\label{app:ablation_nn_milp}

\begin{figure}[ht]
    \centering
    \includegraphics[width=0.9\linewidth]{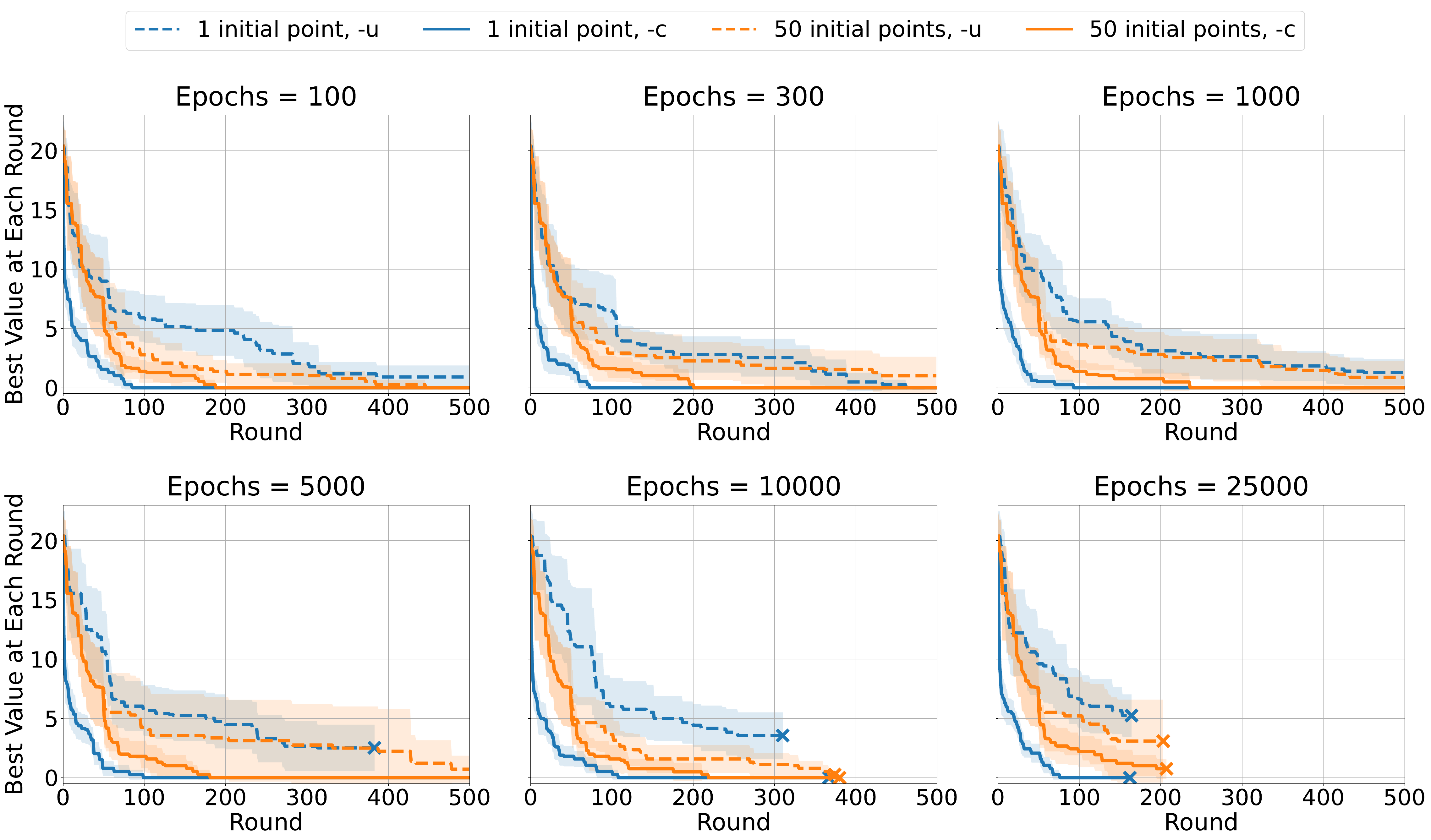}
    \caption{
        Representative ablation study (Ackley $65\times65$ grid) for the NN+MILP baseline. 
        Each panel shows optimization progress for different training epochs. 
        Blue and orange lines denote 1 and 50 initial points, respectively. 
        The results across all tasks confirmed that optimal initial points depend on the constraint setting, and returns diminish after 1000 epochs.
    }
    \label{fig:ablation_nn_milp_rep}
\end{figure}

We conducted an ablation study to optimize the NN+MILP baseline by varying the number of training epochs ($\{100, 300, 1000, 5000, 10000, 25000\}$) and initial points ($\{1, 50\}$). 
While all five benchmarks were evaluated, Figure~\ref{fig:ablation_nn_milp_rep} displays the Ackley $65 \times 65$ grid as a representative case, as it clearly illustrates the observed trends.

Regarding initial points, a distinct preference emerged based on the constraint setting. 
For constrained problems (-c), a single initial point consistently demonstrated superior or competitive performance. 
Conversely, unconstrained problems (-u) significantly benefited from 50 initial points, which led to faster convergence. 
As for training duration, performance gains generally plateaued at 1000 epochs; further training (up to 25000 epochs) yielded diminishing returns relative to the increased computational cost.

Consequently, we adopted the following configurations for all main experiments: 1 initial point with 1000 epochs for constrained settings (-c), and 50 initial points with 1000 epochs for unconstrained settings (-u). 
These settings offer the optimal trade-off between sample efficiency and computational overhead.

\subsection{Effect of Hyperparameters for PROTES}
\label{app:ablation_protes}

\begin{figure}[ht]
    \centering
    \includegraphics[width=0.9\linewidth]{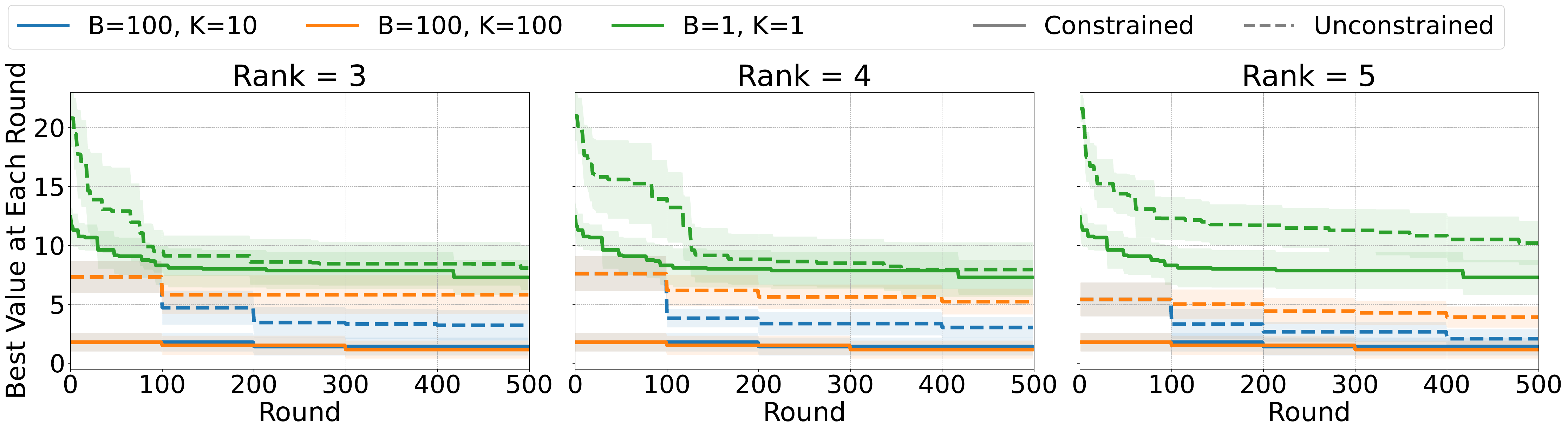}
    \caption{
        Representative ablation study (Ackley $65\times65$ grid) for the PROTES baseline. 
        Curves compare different batch sizes ($B$) and top samples ($K$) across ranks. 
        Consistently across all tasks, $B=100$ and $K=10$ yielded the most stable and effective optimization progress.
    }
    \label{fig:ablation_protes_rep}
\end{figure}

To optimize the PROTES baseline, we evaluated various configurations of batch size ($B \in \{1, 100\}$), top samples ($K \in \{1, 10, 100\}$), and TT-rank ($\{3, 4, 5\}$). 
The results across all tasks consistently showed that a larger batch size of $B=100$ is superior, as it allows the optimizer to gather more comprehensive information about the objective function landscape. 
Figure~\ref{fig:ablation_protes_rep} illustrates these trends using the Ackley $65 \times 65$ grid as a representative example.

Regarding the number of top samples, $K=10$ provided the best balance between focusing on elite samples and maintaining sufficient diversity to avoid premature convergence. 
The performance across TT-ranks 3, 4, and 5 was often similar once optimal $B$ and $K$ values were selected. 
We chose rank 4 as it provides a robust level of model expressiveness without risking unnecessary overfitting.

Based on these consistent observations across all tested tasks, we adopted $B=100$, $K=10$, and rank 4 for all main experiments presented in Section~\ref{sec:experiments}. 
These settings provide the most reliable trade-off between sample efficiency and final optimization performance for the PROTES baseline.

\section{Constraint Violation during Training}
\label{app:violation_check}

\begin{figure}[h]
    \centering
    \includegraphics[width=\linewidth]{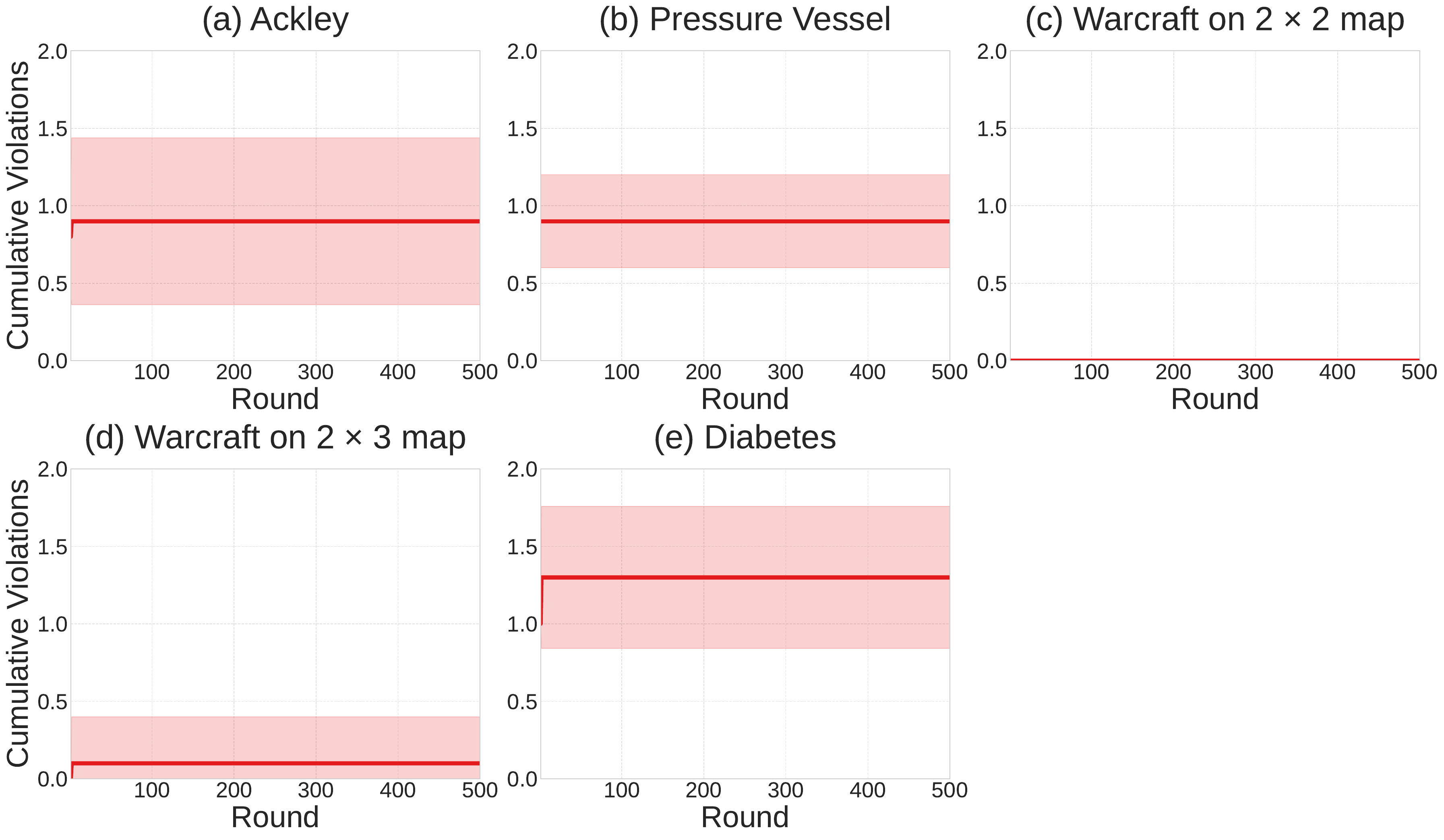}
    \caption{
        Averaged cumulative number of rejected points proposed by CA-TD during SMBO optimization.
    }
    \label{fig:smbo_violation}
\end{figure}

\begin{figure}[h]
    \centering
    \includegraphics[width=\linewidth]{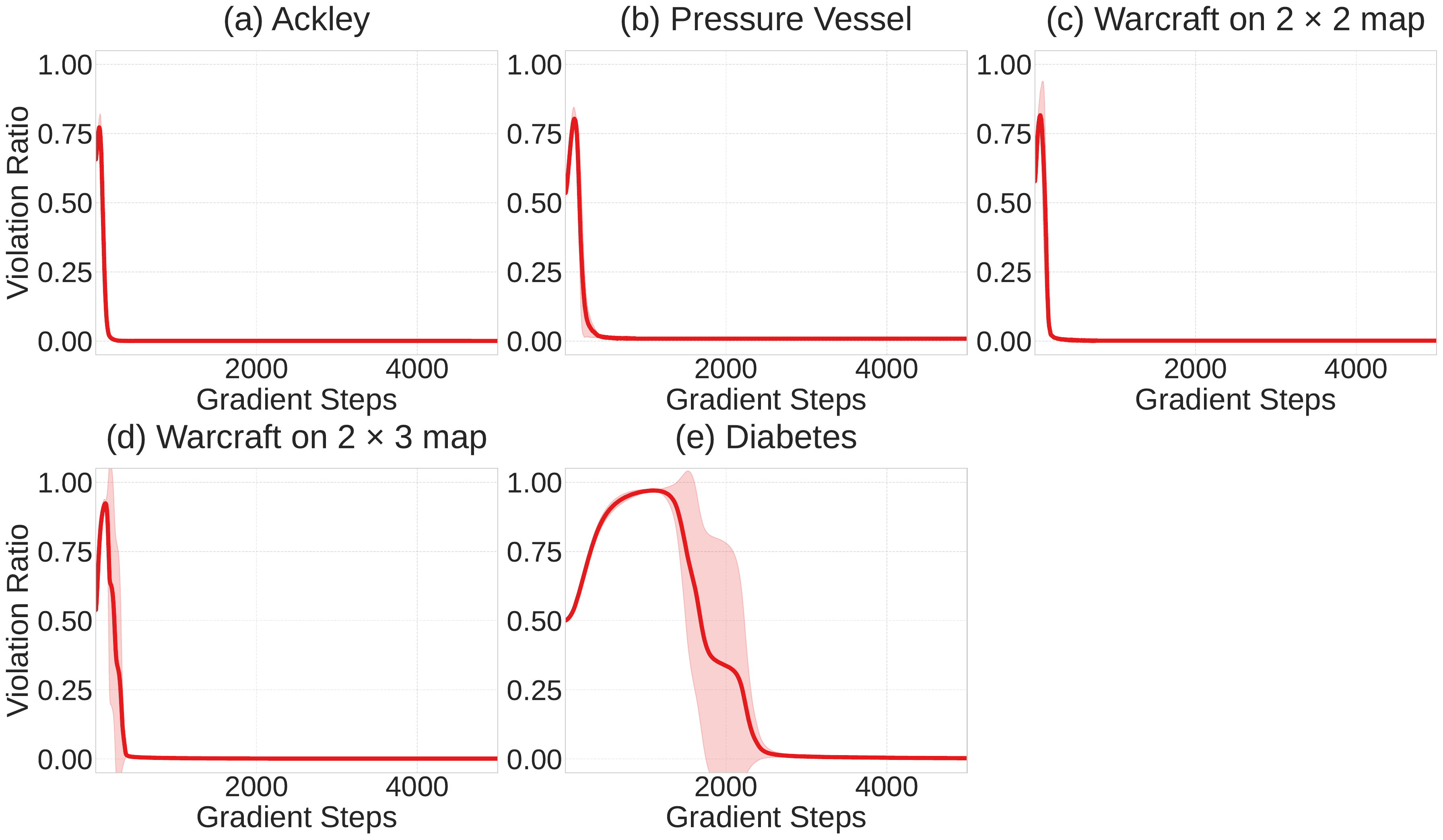}
    \caption{
        Fraction of constraint violations observed during training of the surrogate model in PGRAD. The vertical axis shows the ratio of violated samples to all sampled points in the gradient step.
    }
    \label{fig:pgrad_violation}
\end{figure}

This section reports how frequently constraint violations occur during optimization for both CA-TD and PGRAD.
Figure~\ref{fig:smbo_violation} shows the cumulative averaged number of points proposed by CA-TD during SMBO optimization that were rejected due to constraint violations.
We observe that the number of rejected points does not increase throughout the optimization process, indicating that although CA-TD does not theoretically guarantee feasibility, it practically tends to propose feasible points almost exclusively. 
Figure~\ref{fig:pgrad_violation} presents the fraction of constraint‐violating samples observed during training of the surrogate model when applying PGRAD.
The vertical axis represents the ratio of violated samples to the total number of sampled points. These results are obtained under the setting where no unobserved points remain. The plot demonstrates that PGRAD maintains feasibility for nearly all sampled points during training, with only a very small fraction violating the constraints.
Overall, the empirical evidence shows that CA-TD rarely produces infeasible proposals in practice, and PGRAD is able to prevent constraint violations in almost all cases throughout the training process.

\section{Scalability on High-Dimensional Discrete Spaces}
\label{app:scalability}

\begin{figure}[h]
    \centering
    \includegraphics[width=0.8\linewidth]{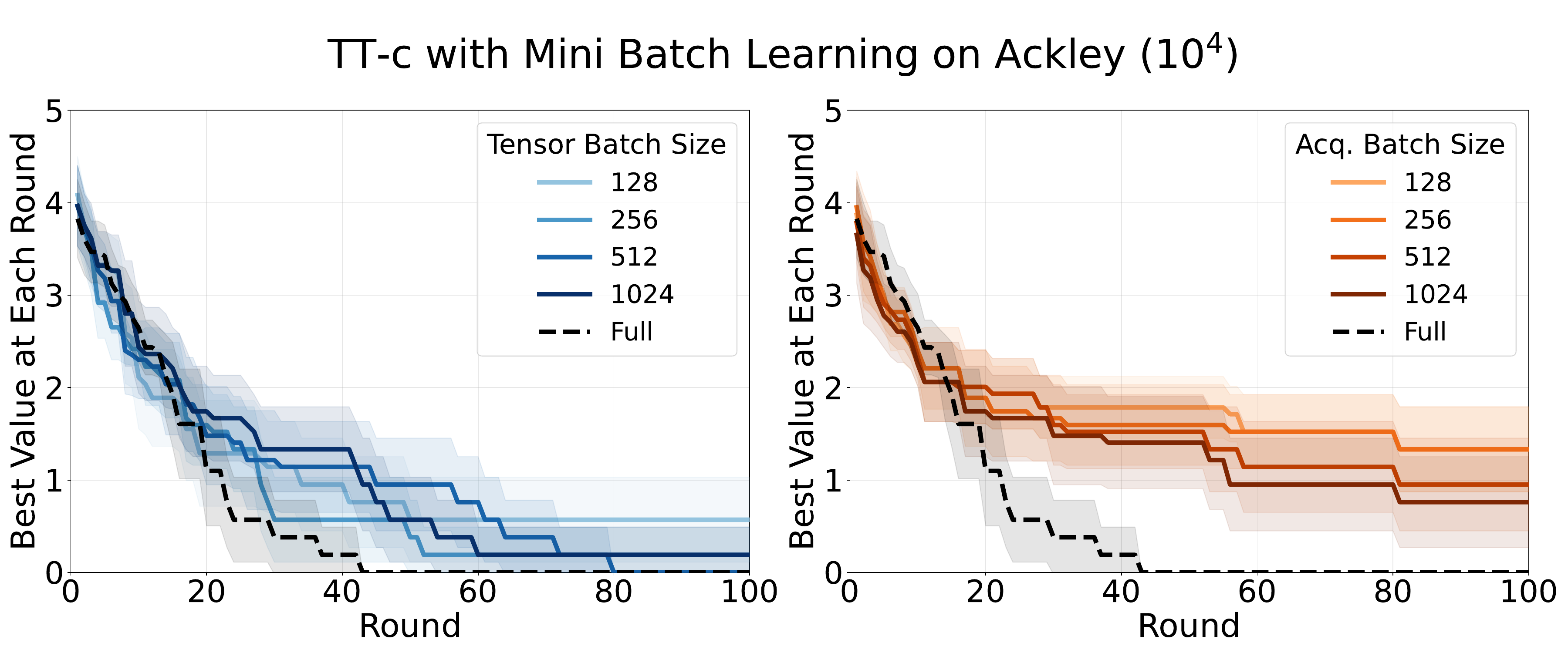} 
    \caption{
        Ablation study on batch size for a search space of size $10^4$. (Left) \textbf{Tensor Batch Size}: Varying the training batch size for PGRAD. (Right) \textbf{Acq. Batch Size}: Varying the inference batch size for acquisition while fixing the training batch size to 128.
    }
    \label{fig:ablation_minibatch_10^4}
\end{figure}

\begin{figure}[h]
    \centering
    \includegraphics[width=0.8\linewidth]{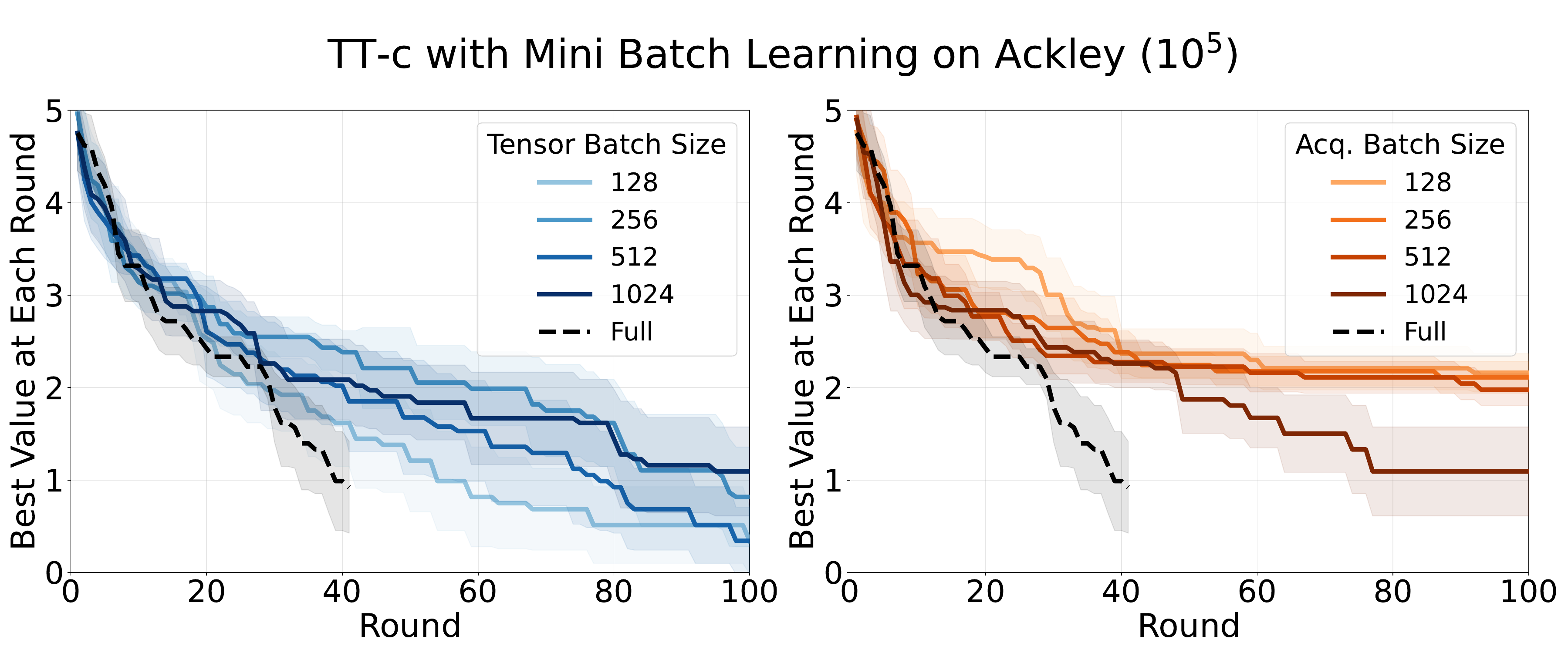} 
    \caption{
        Ablation study on batch size for a search space of size $10^5$. (Left) \textbf{Tensor Batch Size}: Varying the training batch size for PGRAD. (Right) \textbf{Acq. Batch Size}: Varying the inference batch size for acquisition while fixing the training batch size to 128.
    }
    \label{fig:ablation_minibatch_10^5}
\end{figure}

In this section, we present data on the memory usage bottleneck of tensor decomposition-based methods; this is the main bottleneck of our proposed method.
We then demonstrate the impact of batching tensor decomposition, the most naive solution to this bottleneck.

Standard gradient-based optimization methods in deep learning frameworks (e.g., PyTorch) generally require single-precision floating-point (Float32) formats to ensure numerical stability.

Storing dense tensors using single-precision floating-point numbers requires 0.4 GB of memory for a search space size of $10^8$, 4.0 GB for $10^9$, and 40.0 GB for $10^{10}$.

Next, we introduce a mini-batch approach to CA-TD by switching to stochastic optimization.
The mini-batch method used here approximates both the loss function in Eq.~\ref{eq:pen_loss_total} and the acquisition function described in Section~\ref{ssec:ensemble_acquisition} through sampling.
For the loss computation in tensor decomposition, all previously observed points that satisfy the constraints are always included in each batch, while the remaining batch elements are randomly sampled from the constraint-violating points.
For the acquisition function, mini-batches are constructed by uniformly sampling indices from the entire search space.
Furthermore, the gradient descent algorithm is fixed to 200 steps, and at each step only the constraint-violating points are resampled.

This mini-batch method is expected to significantly reduce computational and memory loads by not loading the entire search space into memory.

\subsection{Ablation Study on Batch Size}
\label{app:minibatch}
First, we investigate the impact of mini-batch size selection on optimization.

\paragraph{Experimental Setup }
We utilize the 4-dimensional Ackley function discretized with 10 levels per dimension, resulting in a search space of size $10^4$.
The feasible region is defined by the constraint $\sum_{i=1}^{d} x_i^2 \leq 3^2$.
The computational environment follows the specifications described in Appendix~\ref{app:reproducibility}.
The ``Full'' baseline (black dashed line) represents the ideal setting where both training and acquisition are performed using the full dataset (full-batch).
Larger batches improve convergence toward the full-batch baseline.

\paragraph{Methodology and Results}
We conduct two separate analyses by varying the batch size in ${128, 256, 512, 1024}$, focusing on its effect on (1) tensor training and (2) acquisition inference.

\begin{enumerate}
\item \textbf{Effect of Tensor Batch Size (Surrogate Training):}
We vary the batch size used for the PGRAD updates while keeping the acquisition evaluation in full-batch mode.
As shown in the left panels of Figure~\ref{fig:ablation_minibatch_10^4} and \ref{fig:ablation_minibatch_10^5}, reducing the training batch size substantially degrades performance: smaller batches (e.g., 128) lead to noticeably slower convergence and poorer final objective values. In contrast, larger batches behave similarly to the full-batch baseline, indicating that sufficient tensor batch size is critical for stable surrogate training.
\item \textbf{Effect of Acquisition Batch Size (Inference):}  
We vary the batch size used during the acquisition evaluation while fixing the tensor training batch size to 128.  
The right panels of Figure~\ref{fig:ablation_minibatch_10^4} and \ref{fig:ablation_minibatch_10^5} show that changes in acquisition batch size have almost no effect on the optimization trajectory.
All curves corresponding to different batch sizes closely overlap, indicating that acquisition inference is robust to batch size variation.
\end{enumerate}

\subsection{Applying Mini-Batched CA-TD to High-Dimensional Discrete Space Task}

\begin{figure}[t]
    \centering
    \begin{subfigure}[b]{0.48\textwidth}
        \centering
        \includegraphics[width=\linewidth]{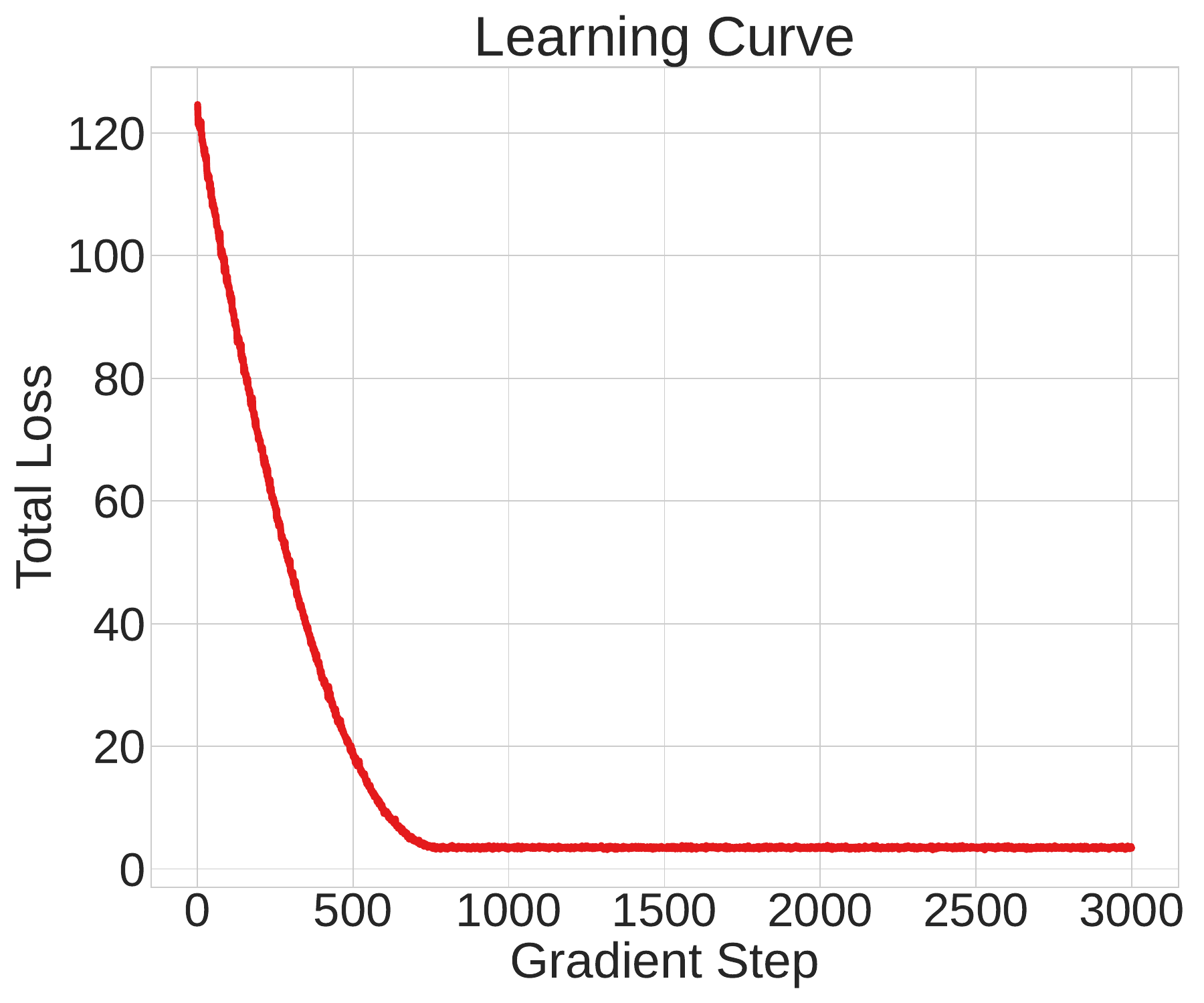}
        \caption{PGRAD training loss ($10^{10}$ Ackley).}
        \label{fig:ackley_large_loss}
    \end{subfigure}
    \hfill
    \begin{subfigure}[b]{0.48\textwidth}
        \centering
        \includegraphics[width=\linewidth]{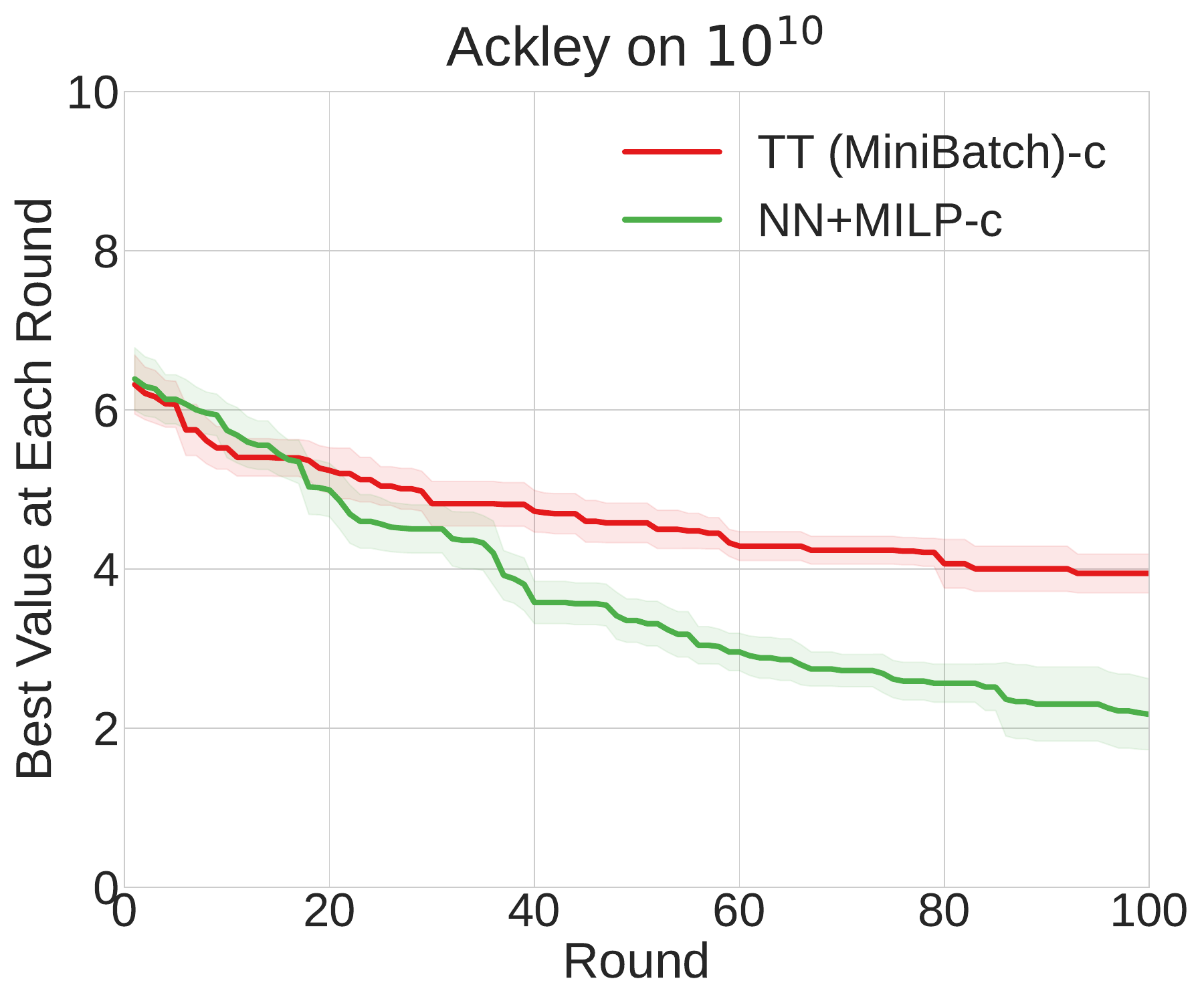}
        \caption{Optimization progress ($10^{10}$ Ackley).}
        \label{fig:ackley_large}
    \end{subfigure}
    \caption{
        Results on the high-dimensional constrained Ackley problem ($10^{10}$ search space). 
        (a) shows the PGRAD convergence with mini-batching. 
        (b) compares CA-TD (TT-c) with the NN+MILP-c baseline.
    }
    \label{fig:high_dim_results}
\end{figure}

Next, we conduct experiments in a search space of size 
$10^{10}$ to demonstrate that the proposed mini-batch method enables optimization even when direct full-batch tensor decomposition is computationally infeasible.

With a mini-batch size of 1024, the execution time per iteration was approximately 4.43 ms, and the convergence behavior is shown in Figure~\ref{fig:ackley_large_loss}.
Figure~\ref{fig:ackley_large} further compares CA-TD with NN+MILP on the same task.
The results show that CA-TD, which leverages constraint information, achieves superior performance in the early optimization phase. In contrast, NN+MILP attains better performance in the later stages.

This performance shift is consistent with the analysis in the previous section: mini-batched CA-TD is sensitive to the batch size used for surrogate training, and in large-scale problems that require highly accurate tensor decompositions, approximation errors accumulate during the later optimization phase. As a result, CA-TD’s effectiveness diminishes over time, whereas NN+MILP maintains stable performance throughout.

\section{Connection to Polynomial Optimization Problems (POPs)}
\label{app:pop}

We clarify why the constrained surrogate learning problem becomes a Polynomial Optimization Problem (POP) by using the TT-based surrogate representation.

The surrogate tensor is represented in Tensor Train (TT) format (\Eqref{eq:tt_entry}).
Expanding the matrix products yields
\[
\hat{\mathcal{Y}}[\mathbf{x}]
=
\sum_{a_0,\dots,a_d}
G^{(1)}_{a_0,a_1}[x_1]\,
G^{(2)}_{a_1,a_2}[x_2]\,
\cdots\,
G^{(d)}_{a_{d-1},a_d}[x_d],
\]
where $a_0,\dots,a_d$ are the indices for summing in the range $1$ to $r_0,\dots,r_d$, respectively ($r_0=r_d=1$).
Thus, $\hat{\mathcal{Y}}[\mathbf{x}]$ is a $d$-degree multivariate polynomial related to the TT core tensor parameters.

The squared error term of the learning objective in \Eqref{eq:pop} is also a polynomial
as $\hat{\mathcal{Y}}[\mathbf{x}]$ is a polynomial in the TT parameters.
Therefore, the entire objective function is a polynomial in the decision variables.

Input constraints impose
$\hat{\mathcal{Y}}[\mathbf{x}] \ge \tau$
for all $\mathbf{x}\in\mathcal{X}_{\mathrm{infeas}}$.
Because $\hat{\mathcal{Y}}[\mathbf{x}]$ is a polynomial, each constraint can be written as
\[
\hat{\mathcal{Y}}[\mathbf{x}] - \tau \ge 0,
\]
which is a polynomial inequality in the TT parameters.

A POP is generally written as
\[
\begin{aligned}
\text{minimize} &\quad f(\theta) \\
\text{subject to} &\quad g_i(\theta) \ge 0, \quad i=1,\dots,m,
\end{aligned}
\]
where both the objective $f$ and constraints $g_i$ are polynomials in the decision variables $\theta$.

The TT-based constrained surrogate learning problem matches this form:
\begin{itemize}
    \item Decision variables $\theta$: all TT core tensor elements $G^{(k)}[x_k]$,
    \item Objective $f(\theta)$: polynomial least-squares error,
    \item Constraints $g_i(\theta)\ge 0$: polynomial lower-bound constraints $\hat{\mathcal{Y}}[\mathbf{x}] - \tau \ge 0$ ($m=|\mathcal{X}_{\mathrm{infeas}}|$).
\end{itemize}

Because both the objective function and all constraints are multivariate polynomials in the TT parameters, the constrained surrogate model learning problem constitutes a POP.

\end{document}